\documentclass[sigconf]{acmart}
\usepackage{graphicx}
\usepackage{tikz}
\usepackage{adjustbox}
\usepackage{soul}
\usepackage{dblfloatfix}
\usepackage{multicol}
\usepackage{enumitem}
\usepackage[caption=false]{subfig}
\usepackage{ragged2e}
\usepackage[toc,page]{appendix}
\usepackage{tcolorbox} % for graphics placeholders

\def\BibTeX{{\rm B\kern-.05em{\sc i\kern-.025em b}\kern-.08emT\kern-.1667em\lower.7ex\hbox{E}\kern-.125emX}}
\copyrightyear{2022}
\acmYear{2022}
\setcopyright{acmlicensed}
\acmConference[ACMGIS'22]{ACM ACM SIGSPATIAL International Conference on Advances in Geographic Information Systems}{November 2022}{Seattle, WA, USA}
\acmPrice{15.00}
\acmDOI{10.1145/1122445.1122456}
\acmISBN{978-1-4503-9999-9/18/06}

\begin{document}

\title{Geo-Adaptive Deep Spatio-Temporal predictive modeling for human mobility}

\author{Syed Mohammed Arshad Zaidi}
\affiliation{%
\institution{University at Buffalo}
\city{Buffalo}
\state{New York}
\country{USA}}
\email{szaidi2@buffalo.edu}

\author{Varun Chandola}
\affiliation{%
\institution{University at Buffalo}
\city{Buffalo}
\state{New York}
\country{USA}}
\email{chandola@buffalo.edu}

\author{EunHye Yoo}
\affiliation{%
\institution{University at Buffalo}
\city{Buffalo}
\state{New York}
\country{USA}}
\email{eunhye@buffalo.edu}

\renewcommand{\shortauthors}{Zaidi et al.}

\begin{abstract}
Deep learning approaches for spatio-temporal prediction problems such as crowd-flow prediction assumes data to be of fixed and regular shaped tensor and face challenges of handling irregular, sparse data tensor.\ This poses limitations in use-case scenarios such as predicting visit counts of individuals' for a given spatial area at a particular temporal resolution using raster/image format representation of the geographical region, since the movement patterns of an individual can be largely restricted and localized to a certain part of the raster.\ Additionally, current deep-learning approaches for solving such problem doesn't account for the geographical awareness of a region while modelling the spatio-temporal movement patterns of an individual.\ To address these limitations, there is a need to develop a novel strategy and modeling approach that can handle both sparse, irregular data while incorporating geo-awareness in the model.\ In this paper, we make use of quadtree as the data structure for representing the image and introduce a novel geo-aware enabled deep learning layer, \textit{GA-ConvLSTM} that performs the convolution operation based on a novel geo-aware module based on quadtree data structure for incorporating spatial dependencies while maintaining the recurrent mechanism for accounting for temporal dependencies.\ We present this approach in the context of the problem of predicting spatial behaviors of an individual (e.g., frequent visits to specific locations) through deep-learning based predictive model, \textit{GADST-Predict}.\ Experimental results on two GPS based trace data shows that the proposed method is effective in handling frequency visits over different use-cases with considerable high accuracy.
\end{abstract}




\maketitle

\section{Introduction}
Mining individuals' mobility patterns is an important research area.\ Studies related to mining mobility patterns have been useful in diverse applications which includes monitoring public health~\cite{alberdi2018smart, barros2018disease}, emergency event detection~\cite{gray2018coupled}, urban planning~\cite{xia2018exploring}, and transportation engineering~\cite{huang2018modeling}.\ Literature ~\cite{song2010modelling} shows that daily human mobility patterns can be captured as a sequence of locations that individuals visited frequently, as they are the basis of essential travel activities such as going from home to workplace in the morning and coming back to home from work in the evening.\ Despite differences among individuals, previous studies ~\cite{cuttone2018understanding, barbosa2018human} have shown that large scale human mobility patterns are highly regular and predictable due to circadian patterns and routine daily activities, such as one's journey to work or home.\ Predicting individuals' future visit counts at different locations can be beneficial to urban planners, policymakers and other entities such as transportation providers and taxi services.

With the rapid increase in use of mobile phones, internet of things, social media platforms across the world, it is now possible to obtain significantly accurate estimate data involving human movements at several temporal and spatial scales.\ Mobility data sources such as Global System for Mobiles (GSM) cell-tower, Global Positioning System (GPS) and Wireless (WiFi) network have started to be used more frequently for mining key insights about individuals' mobility patterns.\ The study~\cite{lin2014mining} provided a brief comparison among data arising from these data sources.\ GPS based data is of higher resolution as compare to other data type as it provide a more accurate estimate of geographical position of mobile devices.\ 

Spatio-temporal prediction problem has been widely studied as it has wide range of applications such as urban planning, traffic control and public health.\ Predictive models for human mobility can help us understand the behavior of migration flow between rural and urban areas~\cite{luca2021leveraging, prieto2018gravity} and the behavioral pattern of movements of individuals' in the presence of natural disasters~\cite{gray2012natural}, climate change and conflicts~\cite{reuveny2007climate}.\ Predictors developed for this task is responsible not only for capturing both spatial and temporal regularities in human movements, but also capture the irregularities in the routine movements.\ In addition to this, the impact of auxiliary factors such as weather impact, use of transportation mode, preferential locations etc.\ on the individual movement patterns can also be captured by such predictors.\ 
Due to high availability of human mobility data with advancements made in the use of deep learning approaches, we have seen unique solutions to predict human movements either at the individual or aggregate level.\ In this study, we focus on the study of predicting an individuals' visit frequency at a finite set of locations using the individual's past mobile phone data.\ This can be seen to be closest variant of crowd-flow prediction task in crowd traffic forecasting where the task is to predict the inflow (number of people entering a region in a given time) and outflow (number of people leaving a region in a given time) of a region given their past history of flow information.\ 

Some studies~\cite{xie2020urban, kamarianakis2005space, kamarianakis2003forecasting} have used traditional multiple machine learning techniques to solve this problem in different settings.\ Deep-learning approaches has been more popularly used modeling approach for tackling this problem over other traditional approaches because it can handle the spatial and temporal dependencies well and can capture the impact of certain external features in increasing the prediction accuracy.\ Deep learning approaches has been used in tackling this task in reference to different applications.\ For example, ~\cite{stec2018forecasting} used deep learning in forecasting crime in a fine-grain city partition, while ~\cite{wang2017deep} used ST-ResNet ~\cite{zhang2017deep} to forecast crime distributions over the Los Angeles area.\ Deep learning approaches have also been used in understanding traffic flow and forecasting traffic accidents.\ For instance,~\cite{yuan2018hetero} used ConvLSTM on heterogeneous urban data for forecasting traffic accidents while~\cite{liu2017short} used ConvLSTM along with Bidirectional LSTM in predicting short-term traffic flow on the urban daily traffic data.\ 
% Crowd counting is another problem in which several deep learning approaches have been employed in the past.\ For example,~\cite{zhang2015cross} used deep convolution neural network to solve the cross-scene crowd problem while use of Bidirectional convLSTM for crowd counting in videos is presented in~\cite{xiong2017spatiotemporal}.\ 
Although, these existing approaches have been employed in various ways on the crowd-flow based problems, they still suffer from the following issues:
\begin{enumerate}
    \item \textbf{Handling large study area with sparsity} -- In general, different studies for crowd-flow based tasks, usually takes the entire area and convert it into raster data format usually into a matrix or tensor.\ This may not be a problem for a small study area, but when the study area gets large, then the ability to handle such large tensors requires a considerable amount of computational resources.\ In addition to this, when there may only small target regions that are used by an individual's movement, then we further cause the algorithm to waste its operation on regions that haven't recorded movements.\ More specifically, we have one individual just having movements recorded near north-east region, while another individual has their movements recorded near south-east.\ Considering these set of individuals with diverse trajectories, the existing approaches will have to perform extra computations to account for spatial and temporal dependencies.
    
    \item \textbf{Incorporating geo-awareness while handling irregularity in geographical data} -- Another issue is that each tensor, which usually has a frequency of visits attached to each of the cell, doesn't account for any geographical information such as presence of water bodies and other hazards, when making the prediction.\ To better enhance the prediction of future count visits in important regions, we can enhance the focus on the regions according to the density of visits made by an individual.\ We make use of varying the coarser and finer resolutions of regions within the geographical area by making use of quad-tree indexing.\ Specifically, the regions with more movements would have finer resolution as compare to the regions with fewer movements.\ This would mean that the specific target regions would end up being irregular gridded which would then pose limitations on the conventional use of the existing deep-learning approaches since most approaches are used to handle fixed and uniform shape of input data.
    
    % \item \textbf{Using robust evaluation metrics} -- Mostly, the approaches used for solving the crowd-flow prediction task are evaluated on the regression-based evaluation metrics, i.e.\ Root Mean Squared Error (RMSE) and mean absolute error (MAE).\ Even though these metrics are widely used for evaluating predictive expressiveness of a model, they can still suffer from vagueness in interpreting results\ As the study area goes from coarser to finer scale, these evaluation measures also gets smaller, which doesn't indicate the correct effectiveness of the model's performance.\ Additionally, when we have a prediction is made in comparison to a very small set of actual ground truth visits concentrated in some small region in the frequency tensor, then we may be more interested in how the model is performing in those regions.
\end{enumerate}

In summary, handling large study area with sparsity, incorporating geo-awareness while handling irregularity in geographical data and using a more robust evaluation metrics are the issues that require great consideration and requires to be addressed accordingly.\ In this paper, we have addressed these issues and have made contributions that can be summarized as follows:
\begin{itemize}
    \item We propose the use of quadtree-based data structure for indexing the regions in a geographical area.\ This is to handle the different density of visit counts across regions, as some regions will have more density of visit counts as compare to other regions.\ Here, we also propose the strategy to employ this data structure in a way so that we incorporate the changes in the architectural model.
    \item We design a novel deep-learning layer called \textit{GA-ConLSTM}, that makes use of geo-awareness module in performing convolutions only in the regions that requires more focus or are more significant in terms of movement patterns of individuals'.
    % \item We make use of a more robust evaluation metrics for evaluating the performance of our proposed model in predicting future visit counts of individuals' for specific quadtree-based index regions.\ These metrics are pretty useful for evaluating the comparisons of the performance of models used for solving task related to crowd-flow prediction.
\end{itemize}

%=================================================================
\section{Background}\label{sec:ch5:mobi2bg}
\subsection{Irregular Geographical data}
\begin{figure}[htbp]
  %\centering
  \subfloat{}{
    \includegraphics[width=0.15\textwidth]{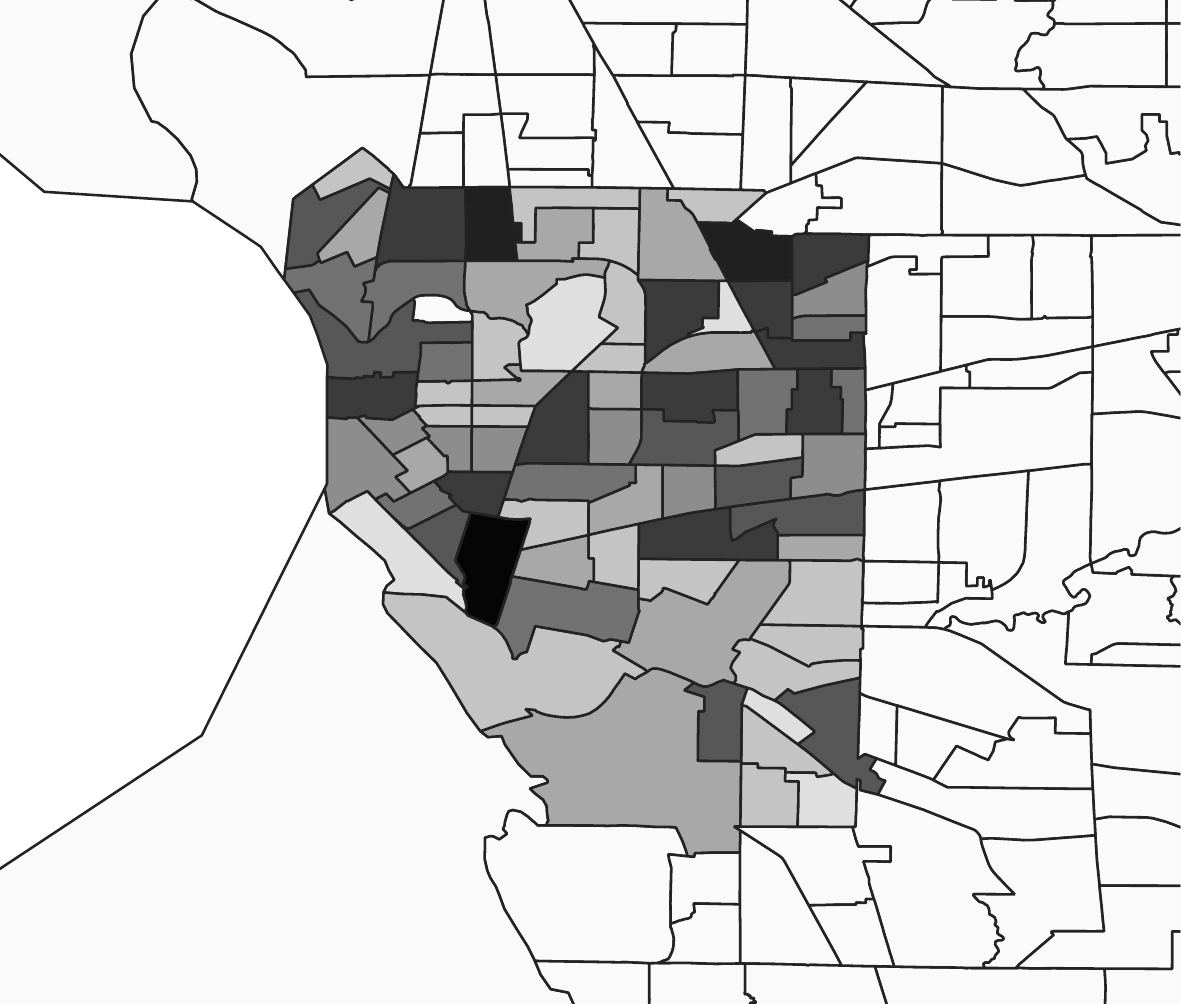}
    \label{fig:irregular1}
  }
  \subfloat{}{
    \includegraphics[width=0.15\textwidth]{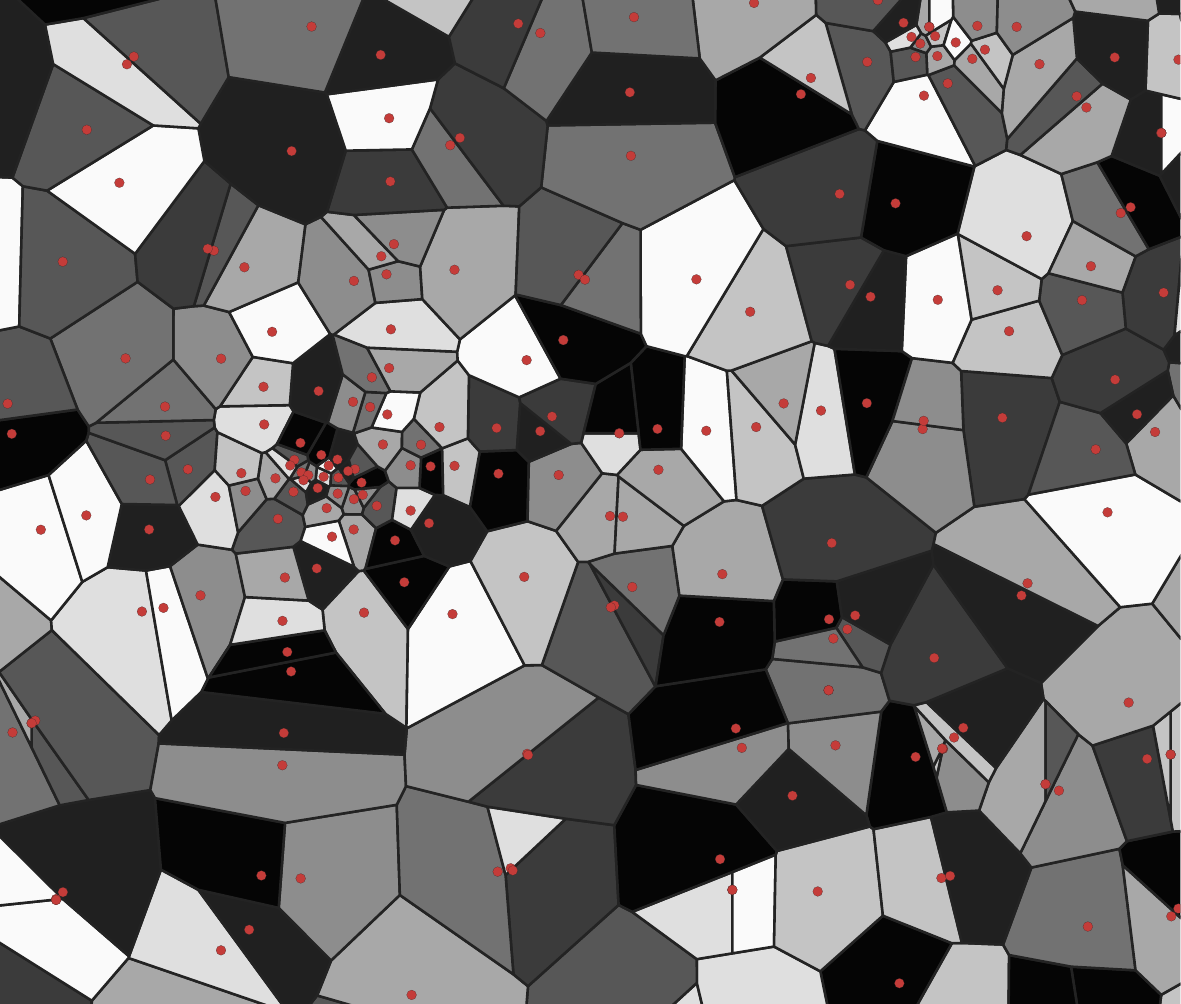}
    \label{fig:irregular2}
  }
  \subfloat{}{
    \includegraphics[width=0.10\textwidth]{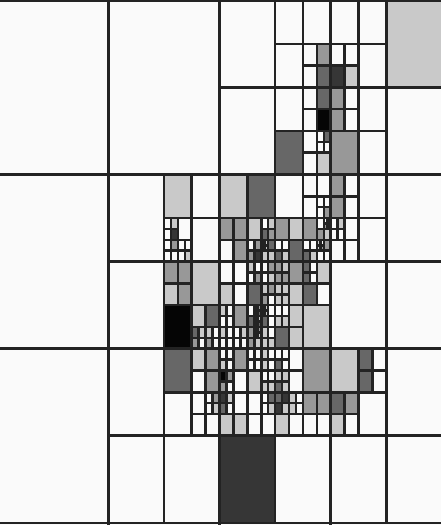}
    \label{fig:irregular3}
  }
  \caption{Three typical examples of irregular ``gridded'' geographic data: a). Each irregular polygon corresponds to an administrative region (census tracts) and the color intensity indicates the number of crime incidents reported within a year, b). Polygons obtained via {\em Voronoi Tesselation} from mobile phone tower locations (shown in red); color intensity indicates the daily number of phone calls originating from each tower, and c). Rectangular polygons obtained from {\em quad-tree} indexing of the target region; color intensity indicates an individual's visit frequency (from GPS traces) to each polygon in a given day.}
  \label{fig:irregulargrids}
\end{figure}

Literature surveys~\cite{atluri2018spatio, wang2019deep} on spatio-temporal data mining provided an overview of taxonomy of different types of data encountered in spatio-temporal problem space.\ These included \textit{event}, \textit{trajectory}, \textit{point reference} and \textit{raster}.\ Different types of problems in this space requires us to represent the data differently as inputs to a model.\ In our work, we are analyzing data in raster or gridded data.\ Raster data are formed as measurements of a continuous or discrete spatio-temporal field recorded at fixed locations in space and time.\ To solve problems related to data involving raster format, we can make use of deep learning approaches like convolutional and recurrent neural network to solve the problems.\ This proved to be beneficial in different applications, however, this has posed certain limitations.\ For example, there could be use-cases where we encounter the sparse data where only a few regions are activated spatially or temporally.\ Due to these challenges, it is imperative to find solutions that effectively handles these limitations.\ In the recent past, many problems have been studied in the context of non-euclidean space that doesn't have restrictions on data to have some form of regular structure.\ For example, a different type of irregular \textit{gridded} geographic data can be seen in Fig.~\ref{fig:irregulargrids}.

%Typical examples of irregular "gridded" geographic data

Employing use of deep learning approaches such as Convolutional Neural Network (CNN) and Recurrent neural network (RNN) for predicting spatio-temporal prediction problem on irregular geographic gridded data is limited, and it requires certain changes either at the algorithm-level or at data-level.\ In this work, we address this problem of building spatio-temporal predictor by handling irregular geographic gridded data based on the rectangular polygons obtained from the quad-tree indexing of the target region, such as shown in Fig.~\ref{fig:irregulargrids}(c).\ This involves proposing novel strategies both at the data-level and algorithmic-level.\ We will first present the proposed methods and will then show the of results on GPS-based trace datasets.

\subsection{Deep Learning for spatio-temporal prediction}
Spatio-temporal data has varying dependencies both at spatial level and temporal level, which poses a considerable limitation to classical data mining methods.\ Deep learning approaches has been widely used in developing predictive models for human mobility~\cite{xiong2017spatiotemporal, yao2018deep, zhang2017deep, yuan2017pred}.\ Predicting future mobility at individual levels is one of the significant problem related to human mobility mining.\ Major prediction tasks concerned with human mobility can be majorly categorized into either \textit{next-location prediction} or \textit{crowd-flow prediction}.\ The problem of next location prediction is concerned with predicting the next location to be visited by the user, given their history of past location's data.\ This task has been found significant in various applications such as improving travel recommendation~\cite{logesh2019exploring, wu2018learning}, geomarketing~\cite{cliquet2020location} and link prediction in social networking platforms~\cite{zheng2018survey}.\ 
%Several deep learning approaches have been proposed for next-location prediction.\ For instance, DeepMove~\cite{feng2018deepmove}, used attention-based recurrent mechanism for mobility prediction from long and sparse trajectories.\ The study~\cite{kong2018hst} proposed HST-LSTM (Hierarchical Spatial-Temporal LSTM) that first used LSTM for combining spatial and temporal characteristics of an individual's trajectory and then extended this to model periodic patterns using encoder-decoder architecture.\
Crowd-flow prediction is concerned with predicting the future incoming and outgoing flows of locations in a geographic region, where the region is tessellated into tiles of shapes such as squares, hexagons, etc.\ This prediction task has significant potential impact for different application use-cases.\ It can help policymakers, development managers, urban and infrastructure planners in deriving key insights from understanding traffic congestion patterns across time.\ This can also benefit certain monitoring agencies in preparing and dealing with adverse situations such as crime and accidents in advance.\ The other benefit can relate to business investment corporations, which can better sense and understand the potential business gains when making investments in a region.\ 
%Traditionally, the use of time-series forecasting methods such as AR, ARMA, ARIMA have been used for this task.\ These methods failed to effectively predict new events since it requires the time-series to be stationary.\ Also, the other limitation of these techniques is that they fail to handle spatial dependencies.\ In order to circumvent these issues, 
Several deep learning based approaches have been widely done in solving this task.\ Most of the studies employing deep-learning based approaches have achieved considerable success with improvements in prediction accuracy.\ Approaches such as Convolution neural networks (CNN) and Recurrent neural networks (RNN) have been used extensively~\cite{clark2017vidloc, zhang2017deep} in capturing spatial and temporal movements in crowd-flow prediction task.\ The study \cite{li2017skeleton} presented the use of both CNN and RNN to capture the spatio-temporal movements.\ Alternatively, a recent study~\cite{xingjian2015convolutional} presented a unique convolution Long-short term network (ConvLSTM) for precipitation nowcasting on radar echo dataset to capture both the spatial and temporal correlation effectively.\ Specifically, some studies have used multiple machine learning techniques to count prediction problems in different settings, including using deep learning approaches to forecast crime incidents across different spatio-temporal scales.\ For example, ~\cite{stec2018forecasting} used deep learning in forecasting crime in a fine-grain city partition, while ~\cite{wang2017deep} used ST-ResNet ~\cite{zhang2017deep} to forecast crime distributions over the Los Angeles area.

% Deep learning approaches have also been used in understanding traffic flow and forecasting traffic accidents.\ For instance,~\cite{yuan2018hetero} used ConvLSTM on heterogeneous urban data for forecasting traffic accidents while~\cite{liu2017short} used ConvLSTM along with Bidirectional LSTM in predicting short-term traffic flow on the urban daily traffic data.\ Crowd counting is another problem in which several deep learning approaches have been employed in the past.\ For example,~\cite{zhang2015cross} used deep convolution neural network to solve the cross-scene crowd problem while use of Bidirectional convLSTM for crowd counting in videos is presented in~\cite{xiong2017spatiotemporal}.\ 
All of these deep-learning based works haven't addressed the problem of incorporating geo-awareness and sparsity of visit counts in the study area explicitly.\ Most of the works have used the study area and converted them into raster format data, which were then fed into the models explicitly.\ In this paper, we propose the use of quadtree data structure to index regions and created a more focused approach where we will provide a model a way to focus on regions that have been frequently visited as compare to other regions.\

%=============================================================================
\section{Problem Formulation}\label{sec:ch5:prob-form}
For each individual, the raw GPS data is available as a series of chronologically ordered GPS locations (latitude and longitude), denoted as $p^{(i)}_1 \rightarrow p^{(i)}_2 \rightarrow \ldots$, where the index in the superscript, $i$, denotes the $i^{th}$ individual.\ 

We transform this data into a gridded representation, by first grouping the locations by a individual-specific temporal window, e.g., hourly, daily, weekly, etc.\ For each window, e.g., a day, we construct a $M \times N$  matrix ${\bf X}^{(i)}_t$, where $M$ and $N$ are the number of rows and columns, respectively, of a uniform spatial grid of a particular scale, applied on the target spatial area. $t$ denotes the index of the temporal window.\ Each entry of ${\bf X}^{(i)}_t$ is equal to the number of times the $i^{th}$ individual's ``visits'' the corresponding grid cell, during the $t^{th}$ window. We will refer to the matrix ${\bf X}^{(i)}_t$  as the {\em visit count matrix} for the $i^{th}$ individual for the $t^{th}$ time window.

We then transform this data into an irregular gridded representation in the form of quadtrees so that we make use of spatial variability of the distribution of visit counts across different regions.\ In this paper, we are investigating the visit frequency prediction problem on quadtrees.\ Here, the visit frequency prediction problem can be defined as follows: {\em Given the historical visit count quadtrees until time $t$, denoted as $\{\mathbf{quadX}^{(i)}_j\}_{j = 1}^t$, predict the future visit counts quadtrees ${\{\bf quadX}^{(i)}_f\}_{f = t+1}^{t+d}$, where $f > t$ and $d$ is the forecasting time steps.} 

%%We bring the spatial variability phenomena to this problem by representing a count matrix as quadtree.\ This is useful in scenarios where we have irregular "gridded" geographical data.

%========================================================================
\section{Methods}\label{sec:ch5:method}
In this section, we will first present Quadtrees.\ Quadtree is a tree-based data structure that can be used to prepare the spatio-temporal raster data.\ We will then introduce the novel convolution recurrent layer that makes use of geo-awareness for performing the convolution while maintaining the recurrent mechanism.\ This ensures that the simultaneous spatial and temporal correlations are considered 

\subsection{Data Structure}
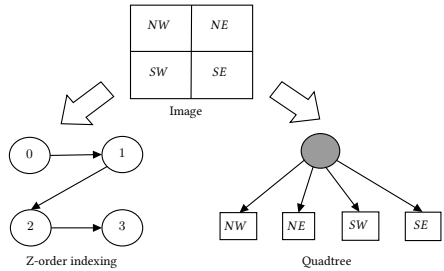
\begin{figure}[htbp] %[x=0.70pt,y=0.75pt,yscale=-1,xscale=1]
\centering

\scalebox{0.55}{
\tikzset{every picture/.style={line width=0.75pt}} %set default line width to 0.75pt        

\begin{tikzpicture}[x=0.70pt,y=0.65pt,yscale=-1,xscale=1]
%uncomment if require: \path (0,642); %set diagram left start at 0, and has height of 642

%Straight Lines [id:da16138617742768524] 
\draw    (290,51) -- (289.5,150) ;
%Shape: Rectangle [id:dp8015639979016513] 
\draw   (230,51) -- (350,51) -- (350,150) -- (230,150) -- cycle ;
%Straight Lines [id:da5140668606249674] 
\draw    (230,100) -- (350,100) ;
%Shape: Ellipse [id:dp8070327066197214] 
\draw  [fill={rgb, 255:red, 155; green, 155; blue, 155 }  ,fill opacity=1 ] (397.71,205.32) .. controls (397.71,195.2) and (406.19,187) .. (416.66,187) .. controls (427.12,187) and (435.61,195.2) .. (435.61,205.32) .. controls (435.61,215.44) and (427.12,223.65) .. (416.66,223.65) .. controls (406.19,223.65) and (397.71,215.44) .. (397.71,205.32) -- cycle ;
%Straight Lines [id:da9372519584432881] 
\draw    (401.12,214.85) -- (339.26,269.02) ;
\draw [shift={(337,271)}, rotate = 318.78999999999996] [fill={rgb, 255:red, 0; green, 0; blue, 0 }  ][line width=0.08]  [draw opacity=0] (8.93,-4.29) -- (0,0) -- (8.93,4.29) -- cycle    ;
%Straight Lines [id:da9932368908561395] 
\draw    (409.46,222.18) -- (395.85,268.12) ;
\draw [shift={(395,271)}, rotate = 286.49] [fill={rgb, 255:red, 0; green, 0; blue, 0 }  ][line width=0.08]  [draw opacity=0] (8.93,-4.29) -- (0,0) -- (8.93,4.29) -- cycle    ;
%Straight Lines [id:da9934779733980383] 
\draw    (421.58,222.18) -- (453.28,267.54) ;
\draw [shift={(455,270)}, rotate = 235.05] [fill={rgb, 255:red, 0; green, 0; blue, 0 }  ][line width=0.08]  [draw opacity=0] (8.93,-4.29) -- (0,0) -- (8.93,4.29) -- cycle    ;
%Straight Lines [id:da20798990900425585] 
\draw    (432.95,215.58) -- (514.48,268.37) ;
\draw [shift={(517,270)}, rotate = 212.92000000000002] [fill={rgb, 255:red, 0; green, 0; blue, 0 }  ][line width=0.08]  [draw opacity=0] (8.93,-4.29) -- (0,0) -- (8.93,4.29) -- cycle    ;
%Shape: Rectangle [id:dp8903301617963619] 
\draw   (318,270.68) -- (354,270.68) -- (354,299.27) -- (318,299.27) -- cycle ;
%Shape: Rectangle [id:dp5727982753982006] 
\draw   (500,270.22) -- (535,270.22) -- (535,300.27) -- (500,300.27) -- cycle ;
%Shape: Rectangle [id:dp9537793792348541] 
\draw   (438,269.95) -- (474,269.95) -- (474,300) -- (438,300) -- cycle ;
%Shape: Rectangle [id:dp7909157307896977] 
\draw   (379.45,270.68) -- (410.91,270.68) -- (410.91,299.27) -- (379.45,299.27) -- cycle ;
%Right Arrow [id:dp4607777974777203] 
\draw   (378.75,132.11) -- (404.55,150.91) -- (410.43,142.82) -- (415.86,171.52) -- (386.88,175.16) -- (392.77,167.07) -- (366.97,148.28) -- cycle ;
%Shape: Ellipse [id:dp7403322350632096] 
\draw   (111,210.13) .. controls (111,199.54) and (119.95,190.96) .. (131,190.96) .. controls (142.05,190.96) and (151,199.54) .. (151,210.13) .. controls (151,220.72) and (142.05,229.31) .. (131,229.31) .. controls (119.95,229.31) and (111,220.72) .. (111,210.13) -- cycle ;
%Shape: Ellipse [id:dp4658618566855084] 
\draw   (202,209.17) .. controls (202,198.58) and (210.95,190) .. (222,190) .. controls (233.05,190) and (242,198.58) .. (242,209.17) .. controls (242,219.76) and (233.05,228.35) .. (222,228.35) .. controls (210.95,228.35) and (202,219.76) .. (202,209.17) -- cycle ;
%Shape: Ellipse [id:dp3222389932397318] 
\draw   (202,286.83) .. controls (202,276.24) and (210.95,267.65) .. (222,267.65) .. controls (233.05,267.65) and (242,276.24) .. (242,286.83) .. controls (242,297.42) and (233.05,306) .. (222,306) .. controls (210.95,306) and (202,297.42) .. (202,286.83) -- cycle ;
%Shape: Ellipse [id:dp5933344004882324] 
\draw   (112,286.83) .. controls (112,276.24) and (120.95,267.65) .. (132,267.65) .. controls (143.05,267.65) and (152,276.24) .. (152,286.83) .. controls (152,297.42) and (143.05,306) .. (132,306) .. controls (120.95,306) and (112,297.42) .. (112,286.83) -- cycle ;
%Straight Lines [id:da5194059915032356] 
\draw    (207.5,221.64) -- (134.56,266.09) ;
\draw [shift={(132,267.65)}, rotate = 328.64] [fill={rgb, 255:red, 0; green, 0; blue, 0 }  ][line width=0.08]  [draw opacity=0] (8.93,-4.29) -- (0,0) -- (8.93,4.29) -- cycle    ;
%Straight Lines [id:da41025405846705154] 
\draw    (152,286.83) -- (199,286.83) ;
\draw [shift={(202,286.83)}, rotate = 180] [fill={rgb, 255:red, 0; green, 0; blue, 0 }  ][line width=0.08]  [draw opacity=0] (8.93,-4.29) -- (0,0) -- (8.93,4.29) -- cycle    ;
%Straight Lines [id:da8799094020804044] 
\draw    (151,210.13) -- (199,209.23) ;
\draw [shift={(202,209.17)}, rotate = 538.9200000000001] [fill={rgb, 255:red, 0; green, 0; blue, 0 }  ][line width=0.08]  [draw opacity=0] (8.93,-4.29) -- (0,0) -- (8.93,4.29) -- cycle    ;
%Right Arrow [id:dp1365076836317647] 
\draw   (213.52,147.8) -- (186.35,169.73) -- (192.63,177.51) -- (161.95,176.56) -- (167.51,146.38) -- (173.79,154.16) -- (200.96,132.24) -- cycle ;

% Text Node
\draw (307,65.4) node [anchor=north west][inner sep=0.75pt]    {$NE$};
% Text Node
\draw (309,115.4) node [anchor=north west][inner sep=0.75pt]    {$SE$};
% Text Node
\draw (247,115.4) node [anchor=north west][inner sep=0.75pt]    {$SW$};
% Text Node
\draw (245,65.4) node [anchor=north west][inner sep=0.75pt]    {$NW$};
% Text Node
\draw (320.39,277.74) node [anchor=north west][inner sep=0.75pt]    {$NW$};
% Text Node
\draw (382.37,278.47) node [anchor=north west][inner sep=0.75pt]    {$NE$};
% Text Node
\draw (443.44,277.21) node [anchor=north west][inner sep=0.75pt]    {$SW$};
% Text Node
\draw (506.96,277.47) node [anchor=north west][inner sep=0.75pt]    {$SE$};
% Text Node
\draw (126,202.53) node [anchor=north west][inner sep=0.75pt]    {$0$};
% Text Node
\draw (217,201.57) node [anchor=north west][inner sep=0.75pt]    {$1$};
% Text Node
\draw (127,278.27) node [anchor=north west][inner sep=0.75pt]    {$2$};
% Text Node
\draw (217,278.27) node [anchor=north west][inner sep=0.75pt]    {$3$};
% Text Node
\draw (267,156) node [anchor=north west][inner sep=0.75pt]   [align=left] {Image};
% Text Node
\draw (397,315.17) node [anchor=north west][inner sep=0.75pt]   [align=left] {Quadtree};
% Text Node
\draw (126,315) node [anchor=north west][inner sep=0.75pt]   [align=left] {Z-order indexing};

\end{tikzpicture}
}
\caption{Example of the quadrants of an image in a quadtree}\label{fig:quadtree}
\end{figure}

In this work, we first decompose an image with the use of linear quadtree ~\cite{gargantini1982effective}.\ Quadtrees ~\cite{finkel1974quad} is a tree data structure where each node has either zero or four children.\ We represent an image with a quadtree where each of non-zero image pixel regions are divided repeatedly into four quadrants (NW -- NorthWest, NE -- NorthEast, SW -- SouthWest, SE -- SouthEast) as shown in Fig. ~\ref{fig:quadtree}

 %Figure -- Quadtree Example

Each of the node in the quadtree is labelled through a \textit{Morton sequencing}~\cite{morton1966computer}, a special sequence that is used as an addressing scheme by interleaving binary digits of $y$ and $x$.\ The decomposition of an image into quadtree leads to the use of morton sequencing address scheme as shown in Fig. ~\ref{fig:blockDec-2}.
%Figure -- Image block decomposition
\input{samples/Figures/combined-blockDecomp-Quad.tex}
Morton sequencing makes use of addressing scheme which is equivalent to interleaving of binary representations of $x$ and $y$ coordinates, and we can also obtain its decimal representation.\ For example, the morton code address of 302 in a quadtree will give the binary representation of row and column coordinates of an image.\ By interleaving the corresponding binary digits of $y = 101$ and $x = 100$, we get 110010 in binary and 50 in decimal).\ We choose the morton code address scheme for location indexing while doing the quadtree decomposition of an image.  

% This address can be used to obtain the z-order index which will be prove to be beneficial when storing the data and index in a contiguous array.\ 

%=================================
\subsection{Data Preparation}
The proposed model consists of first aligning the quadtrees at each level for the raster format data, and then using them as inputs to the proposed deep learning network architecture.\ Here, we will first describe the strategy to align the quadtree and then present the data preparation format that we will need to prepare for the training of our deep-learning based model.

\subsubsection{Quadtree alignment}
The quadtree for each time step or local quadtree would be different, as the visit counts of each user may be different for each day.\ In order to make use of any deep learning approach for temporal modeling, we need to ensure that we align the quadtrees of each timestep in a way that is effective for the use of any deep learning approach.\ 

In order to address the alignment issue, we first construct a universal quadtree by first aligning all the raster image data available in the training set.\ We concatenate all the raster image data and form one universal raster image.\ We then decompose this universal raster image into a universal quadtree.\ Consequently, we will obtain the quadtree that will have the format as a \textit{$<$key, value$>$} where \textit{key} is the quadnode indexed by morton code and \textit{value} is shape of the data indexed by the quadnode index, i.e.\ \textit{key}.\ This information will then be used to align the individual quadtrees so that we have uniformity in the number of regions across different quadtrees.

\subsubsection{Building relation between quadtree levels}
Since the universal quadtree will have nodes at each of the levels, we need to build a relation between nodes at each levels by building a set of nodes we will focus on for preparing the data.\ For example, if we have universal quadtree that has the following \{\textit{key}, \textit{value}\} format which can be equivalently described as \[\{level: [\textit{node index 1}, \textit{node index 2}, \textit{node index 3}, \dots\dots]\}\] 
%Concretely, consider the following example:
%\[\{1:  [0, 1, 2, 3], \quad 2: [00, 01, 02, 03, 10, 11, 12, 13, \dots], \quad 3: [000, 001, 002, 003, \dots], \quad \dots\}\]
where \textit{keys} are the levels, and \textit{values} are the list of quadnodes at each of those level.\ Then, we want a list of quadnodes called \textit{universal quadlist}.\ 

%For example, we can obtain the following universal quadlist.
%\[[0, 100, 101, 102, 103, 20, 21, 22, 23, \dots \dots]\]

This can prove to be useful while preparing the data of each quadtree of a daily raster for each individual in the dataset.\ 

Now, having described the universal quadtrees and universal quadnodes, we can now focus on building the relation between nodes at some current level based on the association between previous and next level.\ One of the ways is to build a relation between all the levels through some parent-child association strength, or how well-connected a parent node is to their child nodes.\ This can be understood through the following example -- if we have quadnode with morton code of 0 in level 1, then we check if the quad node index starting with 0 at level 2 has 4 children nodes or not.\ If we have all 4 children nodes at level 2, i.e.\ 00, 01, 02 and 03, then we include them in the universal quadlist or else we just include the parent node with code 0 at level 1 in the universal quadlist.\ Similarly, if have morton code of 00 at level 2, then we looked into its children at level 3.\ If we have all 4 children nodes at level 3, i.e.\ 000, 001, 002 and 003, then we include them in the universal quadlist or else we stay with the parent node at level 2 and include that in the universal quadlist.\ This way we build a complete list of nodes at different levels.\ 

\subsubsection{Data Transformation}
Once we will have these list of nodes from the universal quadtree, we then perform the alignment of (local) quadtree at each timestep with only the list of quadnodes we got from universal quadtree. 

We align each of the local quadtrees in terms of quadnode indices for each timestep.\ While performing the alignment, there will be some issues.\ These issues can be described as follows:
\begin{itemize}
    \item [--] If a local quadtree at any given timestep doesn't have a quadnode at a level in comparison to the corresponding quadnode in universal quadtree, then we create the quadnode in the local quadtree by creating a data which will be an array of zeros with data shape retrieved for that quadnode from universal quadtree.
    \item [--] If a local quadtree at any given timestep does have a quadnode at a level but is not matching in shape with the quadnode corresponding to the universal quadtree, then we pad the data in the quadnode in the local quadtree with zeros to match the shape with quadnode with universal quadtree.
\end{itemize}

After handling the above issues, we will get an alignment in structure of quadnodes for every local quadtree.\ Next, we transform the data for quadtrees at every timestep as shown in fig.~\ref{fig:dataPrep2}.\ We arrange the quadnodes in a linear fashion and makes use of \{\textit{key}: \textit{value}\} as indicated in fig.~\ref{fig:dataPrep2}.\ We will have a quadnode index (also indicated as qnode\_index) as the key and the list of values where values are quadnode data, index\_start, index\_stop which can be described as below:
\begin{itemize}
    \item \textit{data\_shape}: shape of the data corresponding to quadnode.
    \item \textit{index\_start}: index which indicates the beginning or start position of a given quadnode data array.\ 
    \item \textit{index\_stop}: index which indicates the end or stop position of a given quadnode data array.\ 
\end{itemize}

\input{samples/Figures/ST-Raster2}

%===================================
\subsection{Model Architecture}
Our model comprises the use of a novel geo-aware convolution recurrent layer, called \textit{GA-ConvLSTM} for predicting the future visit frequencies of a user based on its past data.\ Because the data is prepared according to the quadtree-based data structure, it is required to perform both the input and recurrent convolutions based on the adaptive nature of the geographical area.\ This will provide us two major advantages -- \textit{firstly}, we make use of coupled spatio-temporal dependencies that are adaptive to the geographical area and \textit{secondly}, the weights are shared as we go along performing the convolution operation on each of the quad nodes.\ 

\subsubsection{GA-ConvLSTM}
To accommodate both the temporal and spatial dependencies in the data, Shi et al. \cite{xingjian2015convolutional} proposed the Convolutional LSTM (ConvLSTM) that is similar to fully connected LSTM (FC-LSTM) but uses convolution operator in the state-to-state and input-to-state transitions.\ The usual ConvLSTM equations are described below: 
\begin{align*}
    i_t &= \sigma(W_{xi} * x_t + W_{hi} * h_{t-1} + W_{ci} \circ c_{t-1} + b_i)\\
    f_t &= \sigma(W_{xf} * x_t + W_{hf} * h_{t-1} + W_{cf} \circ c_{t-1} + b_f)\\
    c_t &= f_t \circ c_{t-1} + i_t \circ \tanh(W_{xc}x_t + W_{hc}h_{t-1} + b_c)\\
    o_t &= \sigma(W_{xo} * x_t + W_{ho} * h_{t-1} + W_{co} \circ c_t + b_o)\\
    h_t &= o_t \circ \tanh(c_t)
\end{align*}
where $*$ denotes the convolution operation and $\circ$ denotes the Hadamard (elementwise) product.\ Here, $i_t, f_t$ and $o_t$ are the outputs of the input, forget and the output gate respectively. $c_t$ is the cell output at time step, $t$ while $h_t$ is the hidden state of the cell at time step $t$.\ $\sigma$ is the logistic sigmoid function.\ $W_{xi}$, $W_{xf}$, $W_{hi}$, $W_{hf}$, $W_{xc}$, $W_{xo}$, $W_{ho}$, $W_{co}$ corresponds to the weight matrices.\ The usual meaning of each weight parameter matrix is indicative by the subscripts written alongside the symbol ($W$).\ For example, $W_{hi}$ is the weight matrix that maps the hidden to input gate.\ $b_i$, $b_f$, $b_c$, $b_o$ are the bias parameter matrices associated with input gate, forget gate, cell and output gate respectively.

\input{samples/Figures/GA-ConvLSTM}

The conventional ConvLSTM cell takes the input in a fixed and uniform shape, which has to be consistent throughout each of the input timesteps.\ This fixed shape prevents us from modelling any inputs that have varying or irregular shapes.\ In order to accommodate these irregular shapes as inputs, we need to make architectural changes in the convolution operations within the cell.\ This provided us the motivation for proposing fundamental changes into the structure of a conventional ConvLSTM cell, which can be seen as illustrated in figure~\ref{fig:ga-convlstm}.\ We incorporate the use of geo-awareness that directs the convolution operation based on the geographic region.

\subsubsection{Geo-aware module}
We construct a novel geo-awareness module that informs the input and recurrent convolution operations based on the geographical area.\ This awareness module is incorporated through the use of a hash table data structure that has key as quadnode index and value as list which consists of \textit{shape of data within quadnode}, \textit{start\_index } and \textit{stop\_index}.\ The indices i.e.\ \textit{start\_index } and \textit{stop\_index} are meant to handle the extraction of slices of data.\ Once a quadnode data slice is extracted, we then reshape it according to the \textit{data\_shape} value.\ After reshaping this, we then perform the input and recurrent convolutions on this data slice.\ This operation is performed for each of the quadnode data slices.\ The convolution outputs of these data slices are then aggregated together once the operation is performed in all the slices.\ Due to the modifications, our expected tensor shapes for long and short-term memory will also have to be modified.

\subsubsection{Implementation Details}
We first arrange each quadnode and its associated data of a quadtree in a linear sequenced manner.\ For example, if we have the following quadtree with the following quadnodes and quaddata as shown in figure~\ref{fig:quadtree-implement}
\input{samples/Figures/QT-implement}
This way we have a large, flattened set of features.\ The motivation behind building the data like this is to compose the input of shape \textit{(samples, timesteps, features)}.\ We then passed this input to the GA-ConvLSTM layer.\ 

As the input shape is received, we then make use of geo-awareness module for extracting the data in a sliced manner for each of the region indexed by quadnode index.\ Each of the sliced data is reshape into an image format based on the information in \textit{data\_shape} provided by the geo-awareness module.\ Once the data is reshaped, we perform the input and recurrent convolutions on this data slice.\ The convolution operations is done on different data slice similarly.\ Once all the quadnode's data have been convolved, we concatenate all the individual convolution outputs or feature maps together as indicated by \textit{ConvOut} in Fig.\ ~\ref{fig:ga-convlstm} and pass it to the other gates for further processing.\ We also change the expected tensor shapes of long-term memory ($c_t$) and short-term memory ($h_t$) of the usual cell, as we want to ensure that the recurrent mechanism remains unaffected by the proposed modification.

%=============================================================================
\subsection{Proposed Model}
Our proposed model architecture consists of utilizing the novel GA-ConvLSTM layer in predicting future visit counts of individuals, given their past visit counts data.\ The model architecture can be seen in Fig.~\ref{fig:gadst-predict}

\input{samples/Figures/Model3}

Here, we use the proposed GA-ConvLSTM layer to address incorporate geo-awareness while performing spatial convolution operation on irregular gridded geographical data.\ In order to effectively capture any complex spatio-temporal patterns within the visit counts of each individual for one week into the future, we use one week history of visit count observations.\ 

User's location visits are affected by external factors such as type of day (Weekday/Weekend), workplace reported (user is working/not working).\ For example, user's behavior during weekdays will be different from going to some other place during weekends. Similarly, a user's behavior follows some regular pattern going from home to workplace during weekdays while no such regular pattern could be visible during weekends due to irregular spatial and temporal patterns.\ These metadata features aids the model to learn regular time varying changes in the data.\ We handle these features through the external component of the model to further enhance the prediction accuracy of visit counts in the future.\ 

%=======================================
\subsubsection{GA-ConvLSTM based attention-driven encoder-decoder architecture}
The architecture consists of GA-ConvLSTM based attention-driven encoder-decoders that uses past 14 days of historical visit counts data in the quadtree format to forecast the visit counts in the future for 7 days.\ In order to better forecast the daily visit counts upto 1 week in the future, we use the quadtree data representation of the past two weeks in the following manner:
\begin{itemize}
    \item [--] Historical data for week before recent past week, i.e.\ 
    \[{\bf quadX}_{t - 13}, {\bf quadX}_{t - 12}, {\bf quadX}_{t - 11}, \ldots, {\bf quadX}_{t - 7}\]
    \item [--] Historical data for recent past week, i.e.\ 
    \[{\bf quadX}_{t - 6}, {\bf quadX}_{t - 5}, {\bf quadX}_{t - 4}, \ldots, {\bf quadX}_t\]
\end{itemize}

The motivation behind using this is because a simple encoder-decoder scheme may not provide an accurate summary of the past history of observations while we are encoding information as it is restricted to a fixed length of latent vector.\ By incorporating attention mechanism~\cite{bahdanau2014neural}, we account for each of the position of the input sequence while predicting output at each timestep.\ This makes use of the contribution or influence of each data at each position in correspondence with each output.

The working principle of attention mechanism is following: first, we consider if we have $T_x$, number of inputs in the sequence; then, the annotations or hidden state outputs are denoted by ${h}_1, {h}_2, \dots {h}_{T_x}$.\ In the simple encoder-decoder model, only the last state ($h_{T_x}$) of the encoder is used as the context vector and is then passed to the decoder, however, in attention mechanism~\cite{bahdanau2014neural}, we compute the context vector $c_i$ for each target output $y_i$.\ Each of the context vector $c_i$ is generated using a weighted sum of annotations as:
\setcounter{equation}{0}
\begin{eqnarray}
c_i = \sum_{j = 1}^{T_x} \alpha_{ij}h_j
\end{eqnarray}

Here, the weight $\alpha_{ij}$ of each annotation $h_j$ is computed by a softmax function given by the following equation:
\begin{eqnarray}
\alpha_{ij} = \frac{\exp (e_{ij})}{\sum_{k=1}^{T_x} \exp (e_{ik})}
\end{eqnarray}
where 
\begin{eqnarray}
e_{ij} = a(s_{i - 1}, h_j)
\end{eqnarray}

is an alignment model that is responsible for scoring how well the inputs around position $j$ and the output at position $i$ match.\ It is important to note that the score here depends on the hidden state $s_{i - 1}$, which precedes the output $y_i$ and the $j$-th annotation $h_j$ of the input sequence.

\input{samples/Figures/GA-comp2}

In terms of architecture in encoder-decoder LSTM block for this component as shown in Fig.~\ref{fig:GA-seq-to-seq2}, the last ConvLSTM layer is followed by a Batch Normalization, Leaky ReLU activation and a dropout layer before feeding to the attention layer.\ Similarly, on the decoder side, right at the beginning we have a GA-ConvLSTM layer followed by all these layers.\ A rationale behind using these extra layers is to better capture high-level spatial features temporally, which is best used by the attention layer that improves the representation of the past week’s input temporal sequence to generate the relevant output sequence for the following week.
    
\subsubsection{Fusion Layer}
We include a final fusion layer that combines the sequence prediction coming from the two components to predict the final output sequence.\ We compute this output sequence by fusing the sequence outputs of different components of the model with associated learnable component weighted parameters as below:
\begin{equation*}
\hat{\mathbf{Y}} = \mathbf{W}_1 \odot \hat{\mathbf{Y}}_1 + \mathbf{W}_2 \odot \hat{\mathbf{Y}}_2
%+ \mathbf{W}_3 \odot \hat{\mathbf{Y}}_3 + \mathbf{W}_4 \odot \hat{\mathbf{Y}}_4
\end{equation*}
Here, $\hat{\mathbf{Y}}_1$, $\hat{\mathbf{Y}}_2$ are the predicted sequence output coming out of the two components of the model while $\mathbf{W_1}, \mathbf{W_2}$ are the trainable weight parameters that indicate the degree of influence that each of the component has on the final sequence prediction.

\subsubsection{External component}
In order to further enhance the performance of the model, we also use external features concerning individuals' within our data.\ For example, \textit{day of week (weekday/weekend)}, \textit{work reported}, \textit{weather forecast information} etc.\ We make use of the external data concerning the predicted output sequence from the attention-based encoder-decoder component.\ To incorporate the external features such as type of day (weekday/weekend) and individuals' demographics such as gender, age etc.\ as a temporal sequence after the fusion layer, we perform the transformation as shown in Fig.~\ref{fig:gadst-predict}.\ We can represent the external feature sequence as:
\[{\bf ext}_{t+1}, {\bf ext}_{t+2}, \ldots, {\bf ext}_{t + 7}\]

Here, we used two fully connected (Dense) layers wrapped in a Time Distributed layer.\ This first fully connected layer ensures that the dense layer is applied to each element of the input sequence.\ This is followed by a ReLU activation layer, fully connected layer and the subsequent ReLU activation layer.\ Each of these layers are wrapped within a Time Distributed layer.\ The first Dense layer acts as an embedding layer on the external features.\ The second Dense layer is included to map the lower level dimension to the high dimension.\ We then reshape the output to match with the target output coming from the Fusion layer.\ Once the reshape is done, we merge the two outputs to get the final prediction sequence as the output.

\section{Dataset}\label{sec:ch5:datasets}
The proposed approach is validated through the use of two GPS-based trajectory datasets.\ The details of the two datasets are described as below:
\subsection{Dataset 1}
The first data is a private data collected from larger project~\cite{yoo2020quality, yoo2019short, yoo2021impact, eum2021using}.\ A total of 1464 participants who were Apple iPhone users were recruited from October 2016 to June 2017.\ The study area encompasses Buffalo-Niagara region within Erie and Niagara counties of western New York, US. During the study, participants’ locations were collected using their own mobile phone and an application developed by our research team.\ The data has been carefully collected keeping the under the consideration of the privacy of each study participant.\ Primarily, the data set consists of the following information:
\begin{itemize}
    \item \textit{Demographics}: It consists of the participants' personal information such as \textit{gender}, \textit{age group}, \textit{home and work address}, \textit{employment status}.\ In this study, we only use the employment status as an individual-specific feature. In this data set, approximately 17\% of the individuals have a non-working status.
    
    \item \textit{Global Positioning System (GPS) data}: It consists of the movement locations of participants collected at about 35 minute intervals using the application installed on their mobile phones.\ The data was collected for a period of 32 weeks in the years of 2016-2017.
\end{itemize}

% %========================================
\subsection{Dataset 2}
The data is driven from the GeoLife trajectory dataset~\cite{zheng2010geolife}, a publicly available dataset that was collected in GeoLife project by 182 users in a period of over five years (from April 2007 to August 2012).\ It contains 17,621 trajectories with a total distance of about 1.2 million kilometers and a total duration of 48,000+ hours.\ These trajectories were collected using different devices -- GPS receivers and GPS-phones, and have a variety of sampling rates.\ 91 percent of the trajectories are logged in a dense representation, e.g. every $1\sim5$ seconds or every $5\sim10$ meters per point.
Primarily, the data set consists of the following information:
\begin{itemize}
\item \textit{Trajectory data} -- Each point in a GeoLife’s trajectory contains \textit{latitude}, \textit{longitude}, \textit{altitude}, and the \textit{timestamp}.\
\item \textit{Other related information} -- It also contains the user's transportation mode for most of the trajectories.\ This recoded a broad range of users’ outdoor movements, including not only life routines like go home and go to work but also some entertainments and sports activities, such as shopping, sightseeing, dining, hiking, and cycling.
\end{itemize}

%====================================================
\section{Experimental Set-up}
In the data, there were missing data that needed to be handled before proceeding to the training phase.\ This included missing data for days in a sequence for a person.\ We imputed the values of the missing data with the mean value across each corresponding day of the week.\ The observed visit counts at each location were scaled to the range $[0, 1]$.\ While evaluating with the ground truth values, the prediction values are re-scaled back to the normal range.\ The experiments were conducted on a computing cluster available through Centre for Computational Research (CCR) in University at Buffalo.\ The nodes equipped with NVIDIA Tesla V100 GPUs with 16GB memory.\ We used {\em Keras} library~\cite{chollet2015keras} with {\em Tensorflow} library ~\cite{tensorflow2015-whitepaper} as the backend.

\subsection{Model Training}
In Dataset 1, each of 1,464 participants has GPS data records over a maximum period of 221 days (approx.\ 32 weeks), although some participants had less than 221 days.\ On average, participants' GPS data were available on 179 days with a minimum of 53 days.\ Since only 17\% of the 1464 participants had non-working status, we selected data for 10 out of 1464 participants across 32 weeks in a way so that the total participants indicated a well-balanced distribution of working and non-working status.\ We alternatively select participants based on this criteria i.e. out of 485 participants, every alternate participant has a non-working status.\ We train our model using the 60\% of data for each of selected individuals.\ The model was validated with the remaining 40\% data for each selected individuals.\ The description of the data of users used for experimental evaluation is shown in Table~\ref{tab:dataset1}.\ The external features used here is Day of week (Weekday/Weekend) and Work Reported
\begin{table}[htbp]
    %\centering
    \begin{adjustbox}{width=0.45\textwidth} %center=\textwidth
    \begin{tabular}{|c|c|c|c|c|c|c|}
    \hline
    User & Time period & \#days & \#Weekdays & \#Weekends & \#Missing Weekdays & \#Missing Weekends  \\
    \hline
    \hline
    0 &2016/10/24 -- 2017/06/04 &224 &160 &64 &8 &4\\
    1 &2016/10/25 -- 2017/05/21 &209 &149 &60 &5 &8\\
    2 &2016/10/25 -- 2017/06/01 &220 &158 &62 &16 &26\\
    3 &2016/10/25 -- 2017/06/04 &223 &159 &64 &22 &7\\
    4 &2016/10/25 -- 2017/06/04 &223 &159 &64 &25 &11\\
    5 &2016/10/25 -- 2017/05/02 &190 &136 &54 &18 &16\\
    6 &2016/10/25 -- 2017/06/04 &223 &159 &64 &7 &3\\
    7 &2016/10/26 -- 2017/06/04 &222 &158 &64 &6 &4\\
    8 &2016/10/26 -- 2017/05/10 &197 &141 &56 &5 &1\\
    9 &2016/10/26 -- 2017/06/04 &222 &158 &64 &4 &1\\
    10 &2016/10/26 -- 2017/05/22 &209 &149 &60 &11 &5\\
    \hline
    \end{tabular}
    \end{adjustbox}
    \caption{Information about users in Dataset 1}
    \label{tab:dataset1}
\end{table}

In Dataset 2, we first selected the participants with data for most consecutive number of days.\ The objective here was to check whether the model is improving in the absence of any missing data and there is regularity in temporal sequence.\ After the selection, we selected the users with ID 4, 25, 30, 41 and 68.\ The description of the data of users used for experimental evaluation is shown in Table~\ref{tab:dataset2}.\ We train our model using the 60\% of data for each of selected individuals.\ The model was validated with the remaining 40\% data for each selected individuals.\ The external features used here is Day of week (Weekday/Weekend)
\begin{table}[htbp]
    %\centering
    \begin{adjustbox}{width=0.40\textwidth} %center=\textwidth
    \begin{tabular}{|c|c|c|c|c|}
    \hline
    User & Time period & \#days & \#Weekdays & \#Weekends/Holidays  \\
    \hline
    \hline
    4 & 2009/04/01 -- 2009/07/29  & 120 & 84 & 36 \\
    25 & 2009/03/30 -- 2009/06/12 & 75 & 53 & 22 \\ 
    30 & 2009/04/01 -- 2009/07/29 & 120 & 84 & 36 \\ 
    41 & 2009/04/06 -- 2009/07/12 & 98 & 68 & 30 \\ 
    68 & 2009/05/24 -- 2009/08/11 & 80 & 56 & 24 \\ 
    \hline
    \end{tabular}
    \end{adjustbox}
    \caption{Detailed information about users in Dataset 2}
    \label{tab:dataset2}
\end{table}

\vspace{-0.1in}
\subsection{Choosing Hyperparameters}
In the attention-based encoder-decoder part, all the ConvLSTM layers has 40 filters on both the encoder and decoder, with the final ConvLSTM layer in the decoder having 1 filter.\ Each of the filter is of size $3 \times 3$ for extracting the relevant spatial features from both the input and output from the previous timesteps.\ Between each convLSTM layer we have employed a batch normalization layer which is followed by Leaky ReLU and dropout layers.\ The dropout layer is set with the rate of 0.25.\ For the external component, we choose 10 as the number of units for the first Dense layer.
\vspace{-0.1in}
\subsection{Model Training}
We train our model using the training data with batch size of 16 and 300 epochs.\ We used Adam~\cite{kingma2014adam} optimizer with learning rate of 0.001.\ The optimizer is set with $\beta_1 = 0.9$, $\beta_2 = 0.999$, $\epsilon = 1e-07$ and $\text{clip value} = 1.0$.\ We also used model checkpoint that only saves the best weights while training.
\vspace{-0.1in}
\subsection{Loss Function}
{\em Mean squared error} (MSE) is a typical choice for the loss function, in the context of deep learning networks. However, by design, this loss function is biased towards large values, since the square term implies that when there are larger prediction mistakes then the model is punished more than as compare to having smaller prediction mistakes and in order to avoid the prediction being dominated by large values, because of which we also use an additional mean absolute percentage error.\ Loss function, $\mathcal{L}$ for training the model is composed of Mean Square error (MSE) and square of the Mean Absolute Percentage Error (MAPE).

\[\mathcal{L}(\theta) = \frac{\lambda_1}{N} \sum_{i=1}^{N}(Y_i - \hat{Y}_i)^2 + \frac{100 * \lambda_2}{N} * \sum_{i=1}^{N}\left(\frac{Y_i - \hat{Y}_i}{Y_i}\right)^2\]

Here, $\theta$ are all the parameters that needs to be learned in the network and $\lambda_1$ and $\lambda_2$ are the hyperparameters.\ In all our experiments, we choose $\lambda_1 = 10$ and $\lambda_2 = 100$.

%=======================================================
\vspace{-0.1in}
\section{Results and Evaluation}\label{sec:ch5:expResults}
In this section, we will first discuss results of evaluating the model on the given two datasets.\ Subsequently, we will discuss the comparison of our proposed model approach with other competitive baseline approach.\ We use the evaluation metrics, normalized precision (norm\_precision) and normalized recall (norm\_recall) as proposed in our previous study~\cite{zaidi2021dst} and described in the appendix section on evaluation measure.
% Briefly, these evaluation metrics can measure the following two aspects:
% \begin{enumerate}
%     \item {\bf Recall} - {\em What fraction of actual visits were correctly predicted by the model?}
%     \item {\bf Precision} - {\em What fraction of the predicted visits corresponded to the actual visits made by the individual?}
% \end{enumerate}
% Mathematically, the two quantities can be calculated as follows.\ Consider a $(M \times N)$ test image matrix, ${\bf X}$, at a given spatial scale, and let $\widehat{\bf X}$ be the corresponding prediction matrix, obtained from the model. Note that we have dropped the time subscript, $t$, for clarity. The recall and precision are defined as:
% \setcounter{equation}{0}
% \begin{eqnarray}
% \text{recall} & = & \frac{\sum_{i,j=1}^{M,N}\text{min}({\bf X}_{ij},\widehat{{\bf X}}_{ij})}{\sum_{i,j=1}^{M,N}{\bf X}_{ij}}\label{eqn:recall}\\
% \text{precision} & = & \frac{\sum_{i,j=1}^{M,N}\text{min}({\bf X}_{ij},\widehat{{\bf X}}_{ij})}{\sum_{i,j=1}^{M,N}\widehat{\bf X}_{ij}}\label{eqn:precision}
% \end{eqnarray}
% Note that, for both recall and precision, the numerator is the same and counts the overlap between the true and predicted visit counts for each grid cell.\ In the paper study, we report the average recall and precision over all daily visit counts matrices in the test data set.

% %========================================================
\vspace{-0.1in}
\subsection{Results on Dataset 1}
The results for Dataset 1 are summarized in Table~\ref{tab:dataset1-table-rev}.\ Overall, the normalized precision is greater than the normalized recall across different prediction horizons.\ The evaluation measure is consistent as we increase the forecasting horizon from $f = 1$ to $f = 7$. 
%=====================================================================================
%REVISED MODEL (WITH ATTENTION)
%%NEW******* with Precision, Recall, Norm Precision and Norm Recall
%%%%>>>>>>>>>>>>>>>>>>>>>>>>>> REVISED MODEL for 485 users >>>>>>>>>>>>>>>>>>>>>>>>>>>
%%% qt-updateAM-60_40-26x53.out
\begin{table}[htbp]
%\centering
\begin{adjustbox}{width=0.45\textwidth} %{width=1\textwidth}
\small
\begin{tabular}{|c|c|c|c|c|c|c|c|c|} % creating 7 columns
\hline
\textbf{Type} & \textbf{Metric} & \textbf{f = 1}  & \textbf{f = 2} & \textbf{f = 3} &\textbf{f = 4} &\textbf{f = 5} & \textbf{f = 6} & \textbf{f = 7}
\\ [0.5ex]
\hline % inserts single-line

% Entering for 3x3
Overall & norm\_Precision & 0.44 $\pm$ 0.16 & 0.41 $\pm$ 0.17 & 0.42 $\pm$ 0.17 & 0.43 $\pm$ 0.17 & 0.43 $\pm$ 0.18 & 0.45 $\pm$ 0.19 & 0.47 $\pm$ 0.20 \\[0.5ex]
& norm\_Recall & 0.31 $\pm$ 0.26 & 0.32 $\pm$ 0.27 & 0.32 $\pm$ 0.27 & 0.31 $\pm$ 0.27 & 0.30 $\pm$ 0.27 & 0.29 $\pm$ 0.28 & 0.28 $\pm$ 0.27 \\[0.5ex] \cmidrule{1-9}

%Weekdays
Weekdays & norm\_Precision & 0.45 $\pm$ 0.16 & 0.43 $\pm$ 0.17 & 0.44 $\pm$ 0.17 & 0.44 $\pm$ 0.17 & 0.45 $\pm$ 0.18 & 0.48 $\pm$ 0.19 & 0.49 $\pm$ 0.19 \\[0.5ex]
& norm\_Recall & 0.33 $\pm$ 0.26 & 0.33 $\pm$ 0.27 & 0.34 $\pm$ 0.27 & 0.33 $\pm$ 0.27 & 0.31 $\pm$ 0.27 & 0.31 $\pm$ 0.28 & 0.29 $\pm$ 0.27 \\[0.5ex] \cmidrule{1-9}

%Weekends
Weekends & norm\_Precision & 0.40 $\pm$ 0.16 & 0.39 $\pm$ 0.17 & 0.39 $\pm$ 0.17 & 0.39 $\pm$ 0.16 & 0.38 $\pm$ 0.17 & 0.39 $\pm$ 0.18 & 0.42 $\pm$ 0.21 \\[0.5ex]
& norm\_Recall & 0.26 $\pm$ 0.26 & 0.28 $\pm$ 0.27 & 0.28 $\pm$ 0.27 & 0.26 $\pm$ 0.27 & 0.25 $\pm$ 0.26 & 0.24 $\pm$ 0.28 & 0.21 $\pm$ 0.25 \\[0.5ex]
\hline % inserts single-line

\end{tabular}
\end{adjustbox}
\caption{Evaluation of prediction on Dataset 1 for different forecasting horizons (indicated as $\bf f$).\ Each value represents the mean $\pm$ standard deviation.\ \textbf{Overall} means including both weekdays and weekends.
}
\label{tab:dataset1-table-rev}
\end{table}

\vspace{-0.1in}
\subsection{Results on Dataset 2}
The results on the given data are summarized in table~\ref{tab:dataset2-table}.\ We observed that the model is able to predict the future visit counts effectively, although, the model is able to perform much better during weekdays as compare to weekends.\ This can be attributed to the fact that we have more number of weekdays as compare to weekends.
%=====================================================================================
%REVISED MODEL (WITH ATTENTION)
%%NEW******* with Precision, Recall, Norm Precision and Norm Recall
%%%%>>>>>>>>>>>>>>>>>>>>>>>>>> REVISED MODEL for 485 users >>>>>>>>>>>>>>>>>>>>>>>>>>>
\begin{table}[htbp]
%\centering
\begin{adjustbox}{width=0.45\textwidth} %{width=1\textwidth}
\small
\begin{tabular}{|c|c|c|c|c|c|c|c|c|} % creating 7 columns
\hline
\textbf{Type} & \textbf{Metric} & \textbf{f = 1}  & \textbf{f = 2} & \textbf{f = 3} &\textbf{f = 4} &\textbf{f = 5} & \textbf{f = 6} & \textbf{f = 7}
\\ [0.5ex]
\hline % inserts single-line

% Entering for 205x227
Overall & norm\_Precision & 0.48 $\pm$ 0.27 & 0.50 $\pm$ 0.24 & 0.45 $\pm$ 0.23 & 0.47 $\pm$ 0.26 & 0.46 $\pm$ 0.27 & 0.39 $\pm$ 0.23 & 0.31 $\pm$ 0.22 \\[0.5ex]
& norm\_Recall & 0.51 $\pm$ 0.33 & 0.49 $\pm$ 0.31 & 0.52 $\pm$ 0.32 & 0.48 $\pm$ 0.33 & 0.49 $\pm$ 0.34 & 0.50 $\pm$ 0.34 & 0.51 $\pm$ 0.35 \\[0.5ex] \cmidrule{1-9}

%Weekdays
Weekdays &norm\_Precision & 0.49 $\pm$ 0.26 & 0.51 $\pm$ 0.23 & 0.46 $\pm$ 0.23 & 0.51 $\pm$ 0.27 & 0.48 $\pm$ 0.27 & 0.43 $\pm$ 0.25 & 0.33 $\pm$ 0.23 \\[0.5ex] 
&norm\_Recall & 0.58 $\pm$ 0.36 & 0.54 $\pm$ 0.34 & 0.59 $\pm$ 0.35 & 0.56 $\pm$ 0.36 & 0.58 $\pm$ 0.37 & 0.57 $\pm$ 0.36 & 0.60 $\pm$ 0.37 \\[0.5ex] \cmidrule{1-9}

%Weekends
Weekends &norm\_Precision & 0.48 $\pm$ 0.27 & 0.49 $\pm$ 0.24 & 0.43 $\pm$ 0.24 & 0.41 $\pm$ 0.23 & 0.44 $\pm$ 0.26 & 0.34 $\pm$ 0.20 & 0.28 $\pm$ 0.19 \\[0.5ex]
&norm\_Recall & 0.43 $\pm$ 0.26 & 0.43 $\pm$ 0.26 & 0.42 $\pm$ 0.24 & 0.37 $\pm$ 0.23 & 0.38 $\pm$ 0.24 & 0.39 $\pm$ 0.28 & 0.41 $\pm$ 0.28 \\[0.5ex]
\hline % inserts single-line

\end{tabular}
\end{adjustbox}
\caption{Evaluation of prediction on Dataset 2 for different forecasting horizons (indicated as $\bf f$).\ Each value represents the mean $\pm$ standard deviation.\ \textbf{Overall} means including both weekdays and weekends.
}
\label{tab:dataset2-table}
\end{table}
% The prediction results for a continuous time period of 1 week at a region indexed by quadnode $'13'$ for User $17$ is presented in Figure~\ref{fig:Results_GL_1_13}.
%========================================================
\vspace{-0.1in}
\subsection{Comparison to baselines}
Table~\ref{table:compbaseline} presents the performance evaluation of our proposed model with other approaches for spatial resolution of $3 \times 3$ at $\bf f$ = 7 using Dataset 1.\ Here, for each user, 60\% is used as training data and the remaining 40\% is used as validation data.\ The methods used for comparisons are:
% %==============================================================================
%>>>>>>>>>>>Comparison of Proposed Model with other baseline approaches
\begin{table}[htbp]
%\centering
\small
\begin{tabular}{|c | c | c |} % centered columns (3 columns)
\hline %inserts double horizontal lines
Model & Norm. Precision & Norm. Recall \\ [0.2ex] % inserts table
\hline\hline % inserts single horizontal line
ARIMA & 0.14 $\pm$ 0.16 & 0.06 $\pm$ 0.15 \\ [0.2ex]
STResNet~\cite{zhang2017deep} & 0.13 $\pm$ 0.07 & 0.17 $\pm$ 0.09 \\ [0.2ex]
Res-ConvLSTM~\cite{wei2018residual} & 0.26 $\pm$ 0.25 & 0.16 $\pm$ 0.22 \\ [0.2ex] \hline
\textbf{GADST-Predict} & {\bf 0.47 $\pm$ 0.20} & {\bf 0.28 $\pm$ 0.27} \\
[0.2ex] % [1ex] adds vertical space
\hline %inserts single line
\end{tabular}
\caption{Performance evaluation of our proposed model (in \textbf{bold}) in comparison with other approaches.\ Each value represents mean $\pm$ stddev.\ This comparison is done using Dataset 1 at ${\bf f} = 7$.\ } 
%3 x 3 grid size --> 26 x 53
\label{table:compbaseline} % is used to refer this table in the text
\end{table}

% % %==============================================================================
% %>>>>>>>>>>>Comparison of Proposed Model with other baseline approaches
% \begin{table*}[thbp] %[htbp]
% \centering
% \small
% \begin{tabular}{|c | c | c |} % centered columns (3 columns)
% \hline %inserts double horizontal lines
% Model & Norm. Precision & Norm. Recall \\ [0.2ex] % inserts table
% \hline\hline % inserts single horizontal line
% ARIMA & 0.14 $\pm$ 0.16 & 0.06 $\pm$ 0.15 \\ [0.2ex]
% STResNet~\cite{zhang2017deep} & 0.04 $\pm$ 0.02 & 0.005 $\pm$ 0.006 \\ [0.2ex]
% Res-ConvLSTM~\cite{wei2018residual} & 0.19 $\pm$ 0.20 & 0.15 $\pm$ 0.21 \\ [0.2ex] \hline
% \textbf{GADST-Predict} & {\bf 0.34 $\pm$ 0.26} & {\bf 0.29 $\pm$ 0.27} \\
% [0.2ex] % [1ex] adds vertical space
% \hline %inserts single line
% \end{tabular}
% \caption{Performance evaluation of our proposed model (in \textbf{bold}) in comparison with other approaches.\ Each value represents mean $\pm$ stddev.\ This comparison is done using Dataset 1 at ${\bf f} = 7$.} 
% %2 x 2 grid size --> 40 x 80 --> RMS optimizer
% \label{table:compbaseline} % is used to refer this table in the text
% \end{table*}
%======================================================================
\begin{enumerate}
    \item \textbf{ARIMA} -- Autoregressive integrated moving average (ARIMA), also known as Box-Jenkins model, is a popular model used for time-series forecasting.\ It uses the historical time series for predicting future values in the series.
    \item \textbf{STResNet}~\cite{zhang2017deep} -- State-of-the-art approach that makes use of convolutional layers and residual networks for spatio-temporal prediction.
    \item \textbf{Res-ConvLSTM}~\cite{wei2018residual} -- STResNet based variant that makes use of ConvLSTM layers for spatio-temporal predictions.
    % \item \textbf{DST-Predict-Ext} -- Here, we make use of incorporating external meta data such as weekday/weekend and employment status into our model.\ This was used to check whether there is any improvement in our prediction results if we fuse the external features into our model right after the fusion layer in a sequential manner.
    % \item \textbf{DST-Predict-without-Attention} -- Here, we switch off the attention mechanism to check the performance of the model.
    % \item \textbf{GADST-Predict} -- 
\end{enumerate}
The results clearly indicates that our proposed model outperforms the state-of-the-art and competitive baseline approaches in terms of normalized recall and normalized precision.\

% %========================================================
% \subsection{Ablation Study}
% In this section, we investigate the comparison of variants of our model.\ The purpose here is to explore the impact of different architecture or model variants on the prediction accuracy with respect to the evaluation metrics.
% \input{Chapter-5/Tables/ablation2}

% The evaluation of performance of variants of the proposed model is shown in table~\ref{tab:variants} and described as below:

% \begin{enumerate}
%     \item \textbf{Model 1} -- This model uses the attention-based encoder decoder component with external features component.
%     \item \textbf{Model 2} -- This model uses only the attention encoder decoder component.
% \end{enumerate}

%========================================
\section{Conclusions}\label{sec:ch5:conc}
In this work, we proposed novel methods to address the problems for sparsity in large study area, incorporating geo-awareness while handling irregularity in the geographical data and evaluating of the model with respect to more robust metrics.\ We conducted our experiments on a public and private GPS-based trace datasets and validated our models thoroughly.

In the future, we will further enhance our existing deep-learning based approach by incorporating time-aware mechanism that can handle irregular temporal sequence.\ With this integration, spatio-temporal predictors will be more robust for handling challenges of missing data, which is more often encounter in real-world use-case scenarios.

%we introduced a data structure and novel deep-learning based layer, GA-ConvLSTM that incorporates geo-aware mechanism to handle irregular geographical gridded data.\ 

% \clearpage
%\processdelayedfloats   %%%<--- here
\bibliographystyle{ACM-Reference-Format}
\bibliography{sample-base}


\begin{thebibliography}{47}

%%% the Web address.

\ifx \showCODEN    \undefined \def \showCODEN     #1{\unskip}     \fi
\ifx \showDOI      \undefined \def \showDOI       #1{#1}\fi
\ifx \showISBNx    \undefined \def \showISBNx     #1{\unskip}     \fi
\ifx \showISBNxiii \undefined \def \showISBNxiii  #1{\unskip}     \fi
\ifx \showISSN     \undefined \def \showISSN      #1{\unskip}     \fi
\ifx \showLCCN     \undefined \def \showLCCN      #1{\unskip}     \fi
\ifx \shownote     \undefined \def \shownote      #1{#1}          \fi
\ifx \showarticletitle \undefined \def \showarticletitle #1{#1}   \fi
\ifx \showURL      \undefined \def \showURL       {\relax}        \fi
% invisible to TeX
\providecommand\bibfield[2]{#2}
\providecommand\bibinfo[2]{#2}
\providecommand\natexlab[1]{#1}
\providecommand\showeprint[2][]{arXiv:#2}

\bibitem[\protect\citeauthoryear{Abadi, Agarwal, Barham, Brevdo, Chen, Citro,
  Corrado, Davis, Dean, Devin, Ghemawat, Goodfellow, Harp, Irving, Isard, Jia,
  Jozefowicz, Kaiser, Kudlur, Levenberg, Man\'{e}, Monga, Moore, Murray, Olah,
  Schuster, Shlens, Steiner, Sutskever, Talwar, Tucker, Vanhoucke, Vasudevan,
  Vi\'{e}gas, Vinyals, Warden, Wattenberg, Wicke, Yu, and Zheng}{Abadi
  et~al\mbox{.}}{2015}]%
        {tensorflow2015-whitepaper}
\bibfield{author}{\bibinfo{person}{Mart\'{\i}n Abadi}, \bibinfo{person}{Ashish
  Agarwal}, \bibinfo{person}{Paul Barham}, \bibinfo{person}{Eugene Brevdo},
  \bibinfo{person}{Zhifeng Chen}, \bibinfo{person}{Craig Citro},
  \bibinfo{person}{Greg~S. Corrado}, \bibinfo{person}{Andy Davis},
  \bibinfo{person}{Jeffrey Dean}, \bibinfo{person}{Matthieu Devin},
  \bibinfo{person}{Sanjay Ghemawat}, \bibinfo{person}{Ian Goodfellow},
  \bibinfo{person}{Andrew Harp}, \bibinfo{person}{Geoffrey Irving},
  \bibinfo{person}{Michael Isard}, \bibinfo{person}{Yangqing Jia},
  \bibinfo{person}{Rafal Jozefowicz}, \bibinfo{person}{Lukasz Kaiser},
  \bibinfo{person}{Manjunath Kudlur}, \bibinfo{person}{Josh Levenberg},
  \bibinfo{person}{Dan Man\'{e}}, \bibinfo{person}{Rajat Monga},
  \bibinfo{person}{Sherry Moore}, \bibinfo{person}{Derek Murray},
  \bibinfo{person}{Chris Olah}, \bibinfo{person}{Mike Schuster},
  \bibinfo{person}{Jonathon Shlens}, \bibinfo{person}{Benoit Steiner},
  \bibinfo{person}{Ilya Sutskever}, \bibinfo{person}{Kunal Talwar},
  \bibinfo{person}{Paul Tucker}, \bibinfo{person}{Vincent Vanhoucke},
  \bibinfo{person}{Vijay Vasudevan}, \bibinfo{person}{Fernanda Vi\'{e}gas},
  \bibinfo{person}{Oriol Vinyals}, \bibinfo{person}{Pete Warden},
  \bibinfo{person}{Martin Wattenberg}, \bibinfo{person}{Martin Wicke},
  \bibinfo{person}{Yuan Yu}, {and} \bibinfo{person}{Xiaoqiang Zheng}.}
  \bibinfo{year}{2015}\natexlab{}.
\newblock \bibinfo{title}{{TensorFlow}: Large-Scale Machine Learning on
  Heterogeneous Systems}.
\newblock
\newblock
\urldef\tempurl%
\url{http://tensorflow.org/}
\showURL{%
\tempurl}
\newblock
\shownote{Software available from tensorflow.org.}


\bibitem[\protect\citeauthoryear{Alberdi, Weakley, Schmitter-Edgecombe, Cook,
  Aztiria, Basarab, and Barrenechea}{Alberdi et~al\mbox{.}}{2018}]%
        {alberdi2018smart}
\bibfield{author}{\bibinfo{person}{Ane Alberdi}, \bibinfo{person}{Alyssa
  Weakley}, \bibinfo{person}{Maureen Schmitter-Edgecombe},
  \bibinfo{person}{Diane~J Cook}, \bibinfo{person}{Asier Aztiria},
  \bibinfo{person}{Adrian Basarab}, {and} \bibinfo{person}{Maitane
  Barrenechea}.} \bibinfo{year}{2018}\natexlab{}.
\newblock \showarticletitle{Smart home-based prediction of multidomain symptoms
  related to Alzheimer's disease}.
\newblock \bibinfo{journal}{\emph{IEEE journal of biomedical and health
  informatics}} \bibinfo{volume}{22}, \bibinfo{number}{6}
  (\bibinfo{year}{2018}), \bibinfo{pages}{1720--1731}.
\newblock


\bibitem[\protect\citeauthoryear{Atluri, Karpatne, and Kumar}{Atluri
  et~al\mbox{.}}{2018}]%
        {atluri2018spatio}
\bibfield{author}{\bibinfo{person}{Gowtham Atluri}, \bibinfo{person}{Anuj
  Karpatne}, {and} \bibinfo{person}{Vipin Kumar}.}
  \bibinfo{year}{2018}\natexlab{}.
\newblock \showarticletitle{Spatio-temporal data mining: A survey of problems
  and methods}.
\newblock \bibinfo{journal}{\emph{ACM Computing Surveys (CSUR)}}
  \bibinfo{volume}{51}, \bibinfo{number}{4} (\bibinfo{year}{2018}),
  \bibinfo{pages}{1--41}.
\newblock


\bibitem[\protect\citeauthoryear{Bahdanau, Cho, and Bengio}{Bahdanau
  et~al\mbox{.}}{2014}]%
        {bahdanau2014neural}
\bibfield{author}{\bibinfo{person}{Dzmitry Bahdanau},
  \bibinfo{person}{Kyunghyun Cho}, {and} \bibinfo{person}{Yoshua Bengio}.}
  \bibinfo{year}{2014}\natexlab{}.
\newblock \showarticletitle{Neural machine translation by jointly learning to
  align and translate}.
\newblock \bibinfo{journal}{\emph{arXiv preprint arXiv:1409.0473}}
  (\bibinfo{year}{2014}).
\newblock


\bibitem[\protect\citeauthoryear{Barbosa, Barthelemy, Ghoshal, James,
  Lenormand, Louail, Menezes, Ramasco, Simini, and Tomasini}{Barbosa
  et~al\mbox{.}}{2018}]%
        {barbosa2018human}
\bibfield{author}{\bibinfo{person}{Hugo Barbosa}, \bibinfo{person}{Marc
  Barthelemy}, \bibinfo{person}{Gourab Ghoshal}, \bibinfo{person}{Charlotte~R
  James}, \bibinfo{person}{Maxime Lenormand}, \bibinfo{person}{Thomas Louail},
  \bibinfo{person}{Ronaldo Menezes}, \bibinfo{person}{Jos{\'e}~J Ramasco},
  \bibinfo{person}{Filippo Simini}, {and} \bibinfo{person}{Marcello Tomasini}.}
  \bibinfo{year}{2018}\natexlab{}.
\newblock \showarticletitle{Human mobility: Models and applications}.
\newblock \bibinfo{journal}{\emph{Physics Reports}}  \bibinfo{volume}{734}
  (\bibinfo{year}{2018}), \bibinfo{pages}{1--74}.
\newblock


\bibitem[\protect\citeauthoryear{Barros, Duggan, and Rebholz-Schuhmann}{Barros
  et~al\mbox{.}}{2018}]%
        {barros2018disease}
\bibfield{author}{\bibinfo{person}{Joana~M Barros}, \bibinfo{person}{Jim
  Duggan}, {and} \bibinfo{person}{Dietrich Rebholz-Schuhmann}.}
  \bibinfo{year}{2018}\natexlab{}.
\newblock \showarticletitle{Disease mentions in airport and hospital
  geolocations expose dominance of news events for disease concerns}.
\newblock \bibinfo{journal}{\emph{Journal of biomedical semantics}}
  \bibinfo{volume}{9}, \bibinfo{number}{1} (\bibinfo{year}{2018}),
  \bibinfo{pages}{18}.
\newblock


\bibitem[\protect\citeauthoryear{Chollet et~al\mbox{.}}{Chollet
  et~al\mbox{.}}{2015}]%
        {chollet2015keras}
\bibfield{author}{\bibinfo{person}{Fran{\c{c}}ois Chollet} {et~al\mbox{.}}}
  \bibinfo{year}{2015}\natexlab{}.
\newblock \bibinfo{title}{Keras}.
\newblock
\newblock


\bibitem[\protect\citeauthoryear{Clark, Wang, Markham, Trigoni, and Wen}{Clark
  et~al\mbox{.}}{2017}]%
        {clark2017vidloc}
\bibfield{author}{\bibinfo{person}{Ronald Clark}, \bibinfo{person}{Sen Wang},
  \bibinfo{person}{Andrew Markham}, \bibinfo{person}{Niki Trigoni}, {and}
  \bibinfo{person}{Hongkai Wen}.} \bibinfo{year}{2017}\natexlab{}.
\newblock \showarticletitle{Vidloc: A deep spatio-temporal model for 6-dof
  video-clip relocalization}. In \bibinfo{booktitle}{\emph{Proceedings of the
  IEEE Conference on Computer Vision and Pattern Recognition}}.
  \bibinfo{pages}{6856--6864}.
\newblock


\bibitem[\protect\citeauthoryear{Cliquet and Baray}{Cliquet and Baray}{2020}]%
        {cliquet2020location}
\bibfield{author}{\bibinfo{person}{G{\'e}rard Cliquet} {and}
  \bibinfo{person}{J{\'e}r{\^o}me Baray}.} \bibinfo{year}{2020}\natexlab{}.
\newblock \bibinfo{booktitle}{\emph{Location-based Marketing: Geomarketing and
  Geolocation}}.
\newblock \bibinfo{publisher}{John Wiley \& Sons}.
\newblock


\bibitem[\protect\citeauthoryear{Cuttone, Lehmann, and Gonz{\'a}lez}{Cuttone
  et~al\mbox{.}}{2018}]%
        {cuttone2018understanding}
\bibfield{author}{\bibinfo{person}{Andrea Cuttone}, \bibinfo{person}{Sune
  Lehmann}, {and} \bibinfo{person}{Marta~C Gonz{\'a}lez}.}
  \bibinfo{year}{2018}\natexlab{}.
\newblock \showarticletitle{Understanding predictability and exploration in
  human mobility}.
\newblock \bibinfo{journal}{\emph{EPJ Data Science}} \bibinfo{volume}{7},
  \bibinfo{number}{1} (\bibinfo{year}{2018}), \bibinfo{pages}{2}.
\newblock


\bibitem[\protect\citeauthoryear{Eum and Yoo}{Eum and Yoo}{2021}]%
        {eum2021using}
\bibfield{author}{\bibinfo{person}{Youngseob Eum} {and} \bibinfo{person}{EunHye
  Yoo}.} \bibinfo{year}{2021}\natexlab{}.
\newblock \showarticletitle{Using GPS-enabled mobile phones to evaluate the
  associations between human mobility changes and the onset of influenza
  illness}.
\newblock \bibinfo{journal}{\emph{Spatial and Spatio-temporal Epidemiology}}
  (\bibinfo{year}{2021}), \bibinfo{pages}{100458}.
\newblock


\bibitem[\protect\citeauthoryear{Finkel and Bentley}{Finkel and
  Bentley}{1974}]%
        {finkel1974quad}
\bibfield{author}{\bibinfo{person}{Raphael~A. Finkel} {and}
  \bibinfo{person}{Jon~Louis Bentley}.} \bibinfo{year}{1974}\natexlab{}.
\newblock \showarticletitle{Quad trees a data structure for retrieval on
  composite keys}.
\newblock \bibinfo{journal}{\emph{Acta informatica}} \bibinfo{volume}{4},
  \bibinfo{number}{1} (\bibinfo{year}{1974}), \bibinfo{pages}{1--9}.
\newblock


\bibitem[\protect\citeauthoryear{Gargantini}{Gargantini}{1982}]%
        {gargantini1982effective}
\bibfield{author}{\bibinfo{person}{Irene Gargantini}.}
  \bibinfo{year}{1982}\natexlab{}.
\newblock \showarticletitle{An effective way to represent quadtrees}.
\newblock \bibinfo{journal}{\emph{Commun. ACM}} \bibinfo{volume}{25},
  \bibinfo{number}{12} (\bibinfo{year}{1982}), \bibinfo{pages}{905--910}.
\newblock


\bibitem[\protect\citeauthoryear{Gray and Mueller}{Gray and Mueller}{2012}]%
        {gray2012natural}
\bibfield{author}{\bibinfo{person}{Clark~L Gray} {and} \bibinfo{person}{Valerie
  Mueller}.} \bibinfo{year}{2012}\natexlab{}.
\newblock \showarticletitle{Natural disasters and population mobility in
  Bangladesh}.
\newblock \bibinfo{journal}{\emph{Proceedings of the National Academy of
  Sciences}} \bibinfo{volume}{109}, \bibinfo{number}{16}
  (\bibinfo{year}{2012}), \bibinfo{pages}{6000--6005}.
\newblock


\bibitem[\protect\citeauthoryear{Gray, Smolyak, Badirli, and Mohler}{Gray
  et~al\mbox{.}}{2018}]%
        {gray2018coupled}
\bibfield{author}{\bibinfo{person}{Kathryn Gray}, \bibinfo{person}{Daniel
  Smolyak}, \bibinfo{person}{Sarkhan Badirli}, {and} \bibinfo{person}{George
  Mohler}.} \bibinfo{year}{2018}\natexlab{}.
\newblock \showarticletitle{Coupled IGMM-GANs for deep multimodal anomaly
  detection in human mobility data}.
\newblock \bibinfo{journal}{\emph{arXiv preprint arXiv:1809.02728}}
  (\bibinfo{year}{2018}).
\newblock


\bibitem[\protect\citeauthoryear{Huang, Ling, Wang, Zhang, Mao, Lin, and
  Wang}{Huang et~al\mbox{.}}{2018}]%
        {huang2018modeling}
\bibfield{author}{\bibinfo{person}{Zhiren Huang}, \bibinfo{person}{Ximan Ling},
  \bibinfo{person}{Pu Wang}, \bibinfo{person}{Fan Zhang},
  \bibinfo{person}{Yingping Mao}, \bibinfo{person}{Tao Lin}, {and}
  \bibinfo{person}{Fei-Yue Wang}.} \bibinfo{year}{2018}\natexlab{}.
\newblock \showarticletitle{Modeling real-time human mobility based on mobile
  phone and transportation data fusion}.
\newblock \bibinfo{journal}{\emph{Transportation research part C: emerging
  technologies}}  \bibinfo{volume}{96} (\bibinfo{year}{2018}),
  \bibinfo{pages}{251--269}.
\newblock


\bibitem[\protect\citeauthoryear{Kamarianakis and Prastacos}{Kamarianakis and
  Prastacos}{2003}]%
        {kamarianakis2003forecasting}
\bibfield{author}{\bibinfo{person}{Yiannis Kamarianakis} {and}
  \bibinfo{person}{Poulicos Prastacos}.} \bibinfo{year}{2003}\natexlab{}.
\newblock \showarticletitle{Forecasting traffic flow conditions in an urban
  network: Comparison of multivariate and univariate approaches}.
\newblock \bibinfo{journal}{\emph{Transportation Research Record}}
  \bibinfo{volume}{1857}, \bibinfo{number}{1} (\bibinfo{year}{2003}),
  \bibinfo{pages}{74--84}.
\newblock


\bibitem[\protect\citeauthoryear{Kamarianakis and Prastacos}{Kamarianakis and
  Prastacos}{2005}]%
        {kamarianakis2005space}
\bibfield{author}{\bibinfo{person}{Yiannis Kamarianakis} {and}
  \bibinfo{person}{Poulicos Prastacos}.} \bibinfo{year}{2005}\natexlab{}.
\newblock \showarticletitle{Space--time modeling of traffic flow}.
\newblock \bibinfo{journal}{\emph{Computers \& Geosciences}}
  \bibinfo{volume}{31}, \bibinfo{number}{2} (\bibinfo{year}{2005}),
  \bibinfo{pages}{119--133}.
\newblock


\bibitem[\protect\citeauthoryear{Kingma and Ba}{Kingma and Ba}{2014}]%
        {kingma2014adam}
\bibfield{author}{\bibinfo{person}{Diederik~P Kingma} {and}
  \bibinfo{person}{Jimmy Ba}.} \bibinfo{year}{2014}\natexlab{}.
\newblock \showarticletitle{Adam: A method for stochastic optimization}.
\newblock \bibinfo{journal}{\emph{arXiv preprint arXiv:1412.6980}}
  (\bibinfo{year}{2014}).
\newblock


\bibitem[\protect\citeauthoryear{Li, Wang, Wang, Hou, and Li}{Li
  et~al\mbox{.}}{2017}]%
        {li2017skeleton}
\bibfield{author}{\bibinfo{person}{Chuankun Li}, \bibinfo{person}{Pichao Wang},
  \bibinfo{person}{Shuang Wang}, \bibinfo{person}{Yonghong Hou}, {and}
  \bibinfo{person}{Wanqing Li}.} \bibinfo{year}{2017}\natexlab{}.
\newblock \showarticletitle{Skeleton-based action recognition using LSTM and
  CNN}. In \bibinfo{booktitle}{\emph{2017 IEEE International Conference on
  Multimedia \& Expo Workshops (ICMEW)}}. IEEE, \bibinfo{pages}{585--590}.
\newblock


\bibitem[\protect\citeauthoryear{Lin and Hsu}{Lin and Hsu}{2014}]%
        {lin2014mining}
\bibfield{author}{\bibinfo{person}{Miao Lin} {and} \bibinfo{person}{Wen-Jing
  Hsu}.} \bibinfo{year}{2014}\natexlab{}.
\newblock \showarticletitle{Mining GPS data for mobility patterns: A survey}.
\newblock \bibinfo{journal}{\emph{Pervasive and mobile computing}}
  \bibinfo{volume}{12} (\bibinfo{year}{2014}), \bibinfo{pages}{1--16}.
\newblock


\bibitem[\protect\citeauthoryear{Liu, Zheng, Feng, and Chen}{Liu
  et~al\mbox{.}}{2017}]%
        {liu2017short}
\bibfield{author}{\bibinfo{person}{Yipeng Liu}, \bibinfo{person}{Haifeng
  Zheng}, \bibinfo{person}{Xinxin Feng}, {and} \bibinfo{person}{Zhonghui
  Chen}.} \bibinfo{year}{2017}\natexlab{}.
\newblock \showarticletitle{Short-term traffic flow prediction with Conv-LSTM}.
  In \bibinfo{booktitle}{\emph{2017 9th International Conference on Wireless
  Communications and Signal Processing (WCSP)}}. IEEE, \bibinfo{pages}{1--6}.
\newblock


\bibitem[\protect\citeauthoryear{Logesh and Subramaniyaswamy}{Logesh and
  Subramaniyaswamy}{2019}]%
        {logesh2019exploring}
\bibfield{author}{\bibinfo{person}{R Logesh} {and} \bibinfo{person}{V
  Subramaniyaswamy}.} \bibinfo{year}{2019}\natexlab{}.
\newblock \showarticletitle{Exploring hybrid recommender systems for
  personalized travel applications}.
\newblock In \bibinfo{booktitle}{\emph{Cognitive informatics and soft
  computing}}. \bibinfo{publisher}{Springer}, \bibinfo{pages}{535--544}.
\newblock


\bibitem[\protect\citeauthoryear{Luca, Barlacchi, Oliver, and Lepri}{Luca
  et~al\mbox{.}}{2021}]%
        {luca2021leveraging}
\bibfield{author}{\bibinfo{person}{Massimiliano Luca}, \bibinfo{person}{Gianni
  Barlacchi}, \bibinfo{person}{Nuria Oliver}, {and} \bibinfo{person}{Bruno
  Lepri}.} \bibinfo{year}{2021}\natexlab{}.
\newblock \showarticletitle{Leveraging Mobile Phone Data for Migration Flows}.
\newblock \bibinfo{journal}{\emph{arXiv preprint arXiv:2105.14956}}
  (\bibinfo{year}{2021}).
\newblock


\bibitem[\protect\citeauthoryear{Morton}{Morton}{1966}]%
        {morton1966computer}
\bibfield{author}{\bibinfo{person}{Guy~M Morton}.}
  \bibinfo{year}{1966}\natexlab{}.
\newblock \showarticletitle{A computer oriented geodetic data base and a new
  technique in file sequencing}.
\newblock  (\bibinfo{year}{1966}).
\newblock


\bibitem[\protect\citeauthoryear{Prieto~Curiel, Pappalardo, Gabrielli, and
  Bishop}{Prieto~Curiel et~al\mbox{.}}{2018}]%
        {prieto2018gravity}
\bibfield{author}{\bibinfo{person}{Rafael Prieto~Curiel}, \bibinfo{person}{Luca
  Pappalardo}, \bibinfo{person}{Lorenzo Gabrielli}, {and}
  \bibinfo{person}{Steven~Richard Bishop}.} \bibinfo{year}{2018}\natexlab{}.
\newblock \showarticletitle{Gravity and scaling laws of city to city
  migration}.
\newblock \bibinfo{journal}{\emph{PloS one}} \bibinfo{volume}{13},
  \bibinfo{number}{7} (\bibinfo{year}{2018}), \bibinfo{pages}{e0199892}.
\newblock


\bibitem[\protect\citeauthoryear{Reuveny}{Reuveny}{2007}]%
        {reuveny2007climate}
\bibfield{author}{\bibinfo{person}{Rafael Reuveny}.}
  \bibinfo{year}{2007}\natexlab{}.
\newblock \showarticletitle{Climate change-induced migration and violent
  conflict}.
\newblock \bibinfo{journal}{\emph{Political geography}} \bibinfo{volume}{26},
  \bibinfo{number}{6} (\bibinfo{year}{2007}), \bibinfo{pages}{656--673}.
\newblock


\bibitem[\protect\citeauthoryear{Song, Koren, Wang, and Barab{\'a}si}{Song
  et~al\mbox{.}}{2010}]%
        {song2010modelling}
\bibfield{author}{\bibinfo{person}{Chaoming Song}, \bibinfo{person}{Tal Koren},
  \bibinfo{person}{Pu Wang}, {and} \bibinfo{person}{Albert-L{\'a}szl{\'o}
  Barab{\'a}si}.} \bibinfo{year}{2010}\natexlab{}.
\newblock \showarticletitle{Modelling the scaling properties of human
  mobility}.
\newblock \bibinfo{journal}{\emph{Nature Physics}} \bibinfo{volume}{6},
  \bibinfo{number}{10} (\bibinfo{year}{2010}), \bibinfo{pages}{818}.
\newblock


\bibitem[\protect\citeauthoryear{Stec and Klabjan}{Stec and Klabjan}{2018}]%
        {stec2018forecasting}
\bibfield{author}{\bibinfo{person}{Alexander Stec} {and} \bibinfo{person}{Diego
  Klabjan}.} \bibinfo{year}{2018}\natexlab{}.
\newblock \showarticletitle{Forecasting Crime with Deep Learning}.
\newblock \bibinfo{journal}{\emph{arXiv preprint arXiv:1806.01486}}
  (\bibinfo{year}{2018}).
\newblock


\bibitem[\protect\citeauthoryear{Wang, Zhang, Zhang, Brantingham, and
  Bertozzi}{Wang et~al\mbox{.}}{2017}]%
        {wang2017deep}
\bibfield{author}{\bibinfo{person}{Bao Wang}, \bibinfo{person}{Duo Zhang},
  \bibinfo{person}{Duanhao Zhang}, \bibinfo{person}{P~Jeffery Brantingham},
  {and} \bibinfo{person}{Andrea~L Bertozzi}.} \bibinfo{year}{2017}\natexlab{}.
\newblock \showarticletitle{Deep learning for real time crime forecasting}.
\newblock \bibinfo{journal}{\emph{arXiv preprint arXiv:1707.03340}}
  (\bibinfo{year}{2017}).
\newblock


\bibitem[\protect\citeauthoryear{Wang, Cao, and Yu}{Wang et~al\mbox{.}}{2019}]%
        {wang2019deep}
\bibfield{author}{\bibinfo{person}{Senzhang Wang}, \bibinfo{person}{Jiannong
  Cao}, {and} \bibinfo{person}{Philip~S Yu}.} \bibinfo{year}{2019}\natexlab{}.
\newblock \showarticletitle{Deep learning for spatio-temporal data mining: A
  survey}.
\newblock \bibinfo{journal}{\emph{arXiv preprint arXiv:1906.04928}}
  (\bibinfo{year}{2019}).
\newblock


\bibitem[\protect\citeauthoryear{Wei, Zhou, Sankaranarayanan, Sengupta, and
  Samet}{Wei et~al\mbox{.}}{2018}]%
        {wei2018residual}
\bibfield{author}{\bibinfo{person}{Hong Wei}, \bibinfo{person}{Hao Zhou},
  \bibinfo{person}{Jangan Sankaranarayanan}, \bibinfo{person}{Sudipta
  Sengupta}, {and} \bibinfo{person}{Hanan Samet}.}
  \bibinfo{year}{2018}\natexlab{}.
\newblock \showarticletitle{Residual convolutional lstm for tweet count
  prediction}. In \bibinfo{booktitle}{\emph{Companion Proceedings of the The
  Web Conference 2018}}. International World Wide Web Conferences Steering
  Committee, \bibinfo{pages}{1309--1316}.
\newblock


\bibitem[\protect\citeauthoryear{Wu, Luo, Yang, and Shao}{Wu
  et~al\mbox{.}}{2018}]%
        {wu2018learning}
\bibfield{author}{\bibinfo{person}{Ruizhi Wu}, \bibinfo{person}{Guangchun Luo},
  \bibinfo{person}{Qinli Yang}, {and} \bibinfo{person}{Junming Shao}.}
  \bibinfo{year}{2018}\natexlab{}.
\newblock \showarticletitle{Learning individual moving preference and social
  interaction for location prediction}.
\newblock \bibinfo{journal}{\emph{IEEE Access}}  \bibinfo{volume}{6}
  (\bibinfo{year}{2018}), \bibinfo{pages}{10675--10687}.
\newblock


\bibitem[\protect\citeauthoryear{Xia, Wang, Kong, Wang, Li, and Liu}{Xia
  et~al\mbox{.}}{2018}]%
        {xia2018exploring}
\bibfield{author}{\bibinfo{person}{Feng Xia}, \bibinfo{person}{Jinzhong Wang},
  \bibinfo{person}{Xiangjie Kong}, \bibinfo{person}{Zhibo Wang},
  \bibinfo{person}{Jianxin Li}, {and} \bibinfo{person}{Chengfei Liu}.}
  \bibinfo{year}{2018}\natexlab{}.
\newblock \showarticletitle{Exploring human mobility patterns in urban
  scenarios: A trajectory data perspective}.
\newblock \bibinfo{journal}{\emph{IEEE Communications Magazine}}
  \bibinfo{volume}{56}, \bibinfo{number}{3} (\bibinfo{year}{2018}),
  \bibinfo{pages}{142--149}.
\newblock


\bibitem[\protect\citeauthoryear{Xie, Li, Liu, Du, Yang, and Zhang}{Xie
  et~al\mbox{.}}{2020}]%
        {xie2020urban}
\bibfield{author}{\bibinfo{person}{Peng Xie}, \bibinfo{person}{Tianrui Li},
  \bibinfo{person}{Jia Liu}, \bibinfo{person}{Shengdong Du},
  \bibinfo{person}{Xin Yang}, {and} \bibinfo{person}{Junbo Zhang}.}
  \bibinfo{year}{2020}\natexlab{}.
\newblock \showarticletitle{Urban flow prediction from spatiotemporal data
  using machine learning: A survey}.
\newblock \bibinfo{journal}{\emph{Information Fusion}}  \bibinfo{volume}{59}
  (\bibinfo{year}{2020}), \bibinfo{pages}{1--12}.
\newblock


\bibitem[\protect\citeauthoryear{Xingjian, Chen, Wang, Yeung, Wong, and
  Woo}{Xingjian et~al\mbox{.}}{2015}]%
        {xingjian2015convolutional}
\bibfield{author}{\bibinfo{person}{SHI Xingjian}, \bibinfo{person}{Zhourong
  Chen}, \bibinfo{person}{Hao Wang}, \bibinfo{person}{Dit-Yan Yeung},
  \bibinfo{person}{Wai-Kin Wong}, {and} \bibinfo{person}{Wang-chun Woo}.}
  \bibinfo{year}{2015}\natexlab{}.
\newblock \showarticletitle{Convolutional LSTM network: A machine learning
  approach for precipitation nowcasting}. In \bibinfo{booktitle}{\emph{Advances
  in neural information processing systems}}. \bibinfo{pages}{802--810}.
\newblock


\bibitem[\protect\citeauthoryear{Xiong, Shi, and Yeung}{Xiong
  et~al\mbox{.}}{2017}]%
        {xiong2017spatiotemporal}
\bibfield{author}{\bibinfo{person}{Feng Xiong}, \bibinfo{person}{Xingjian Shi},
  {and} \bibinfo{person}{Dit-Yan Yeung}.} \bibinfo{year}{2017}\natexlab{}.
\newblock \showarticletitle{Spatiotemporal modeling for crowd counting in
  videos}. In \bibinfo{booktitle}{\emph{Proceedings of the IEEE International
  Conference on Computer Vision}}. \bibinfo{pages}{5151--5159}.
\newblock


\bibitem[\protect\citeauthoryear{Yao, Wu, Ke, Tang, Jia, Lu, Gong, Ye, and
  Li}{Yao et~al\mbox{.}}{2018}]%
        {yao2018deep}
\bibfield{author}{\bibinfo{person}{Huaxiu Yao}, \bibinfo{person}{Fei Wu},
  \bibinfo{person}{Jintao Ke}, \bibinfo{person}{Xianfeng Tang},
  \bibinfo{person}{Yitian Jia}, \bibinfo{person}{Siyu Lu},
  \bibinfo{person}{Pinghua Gong}, \bibinfo{person}{Jieping Ye}, {and}
  \bibinfo{person}{Zhenhui Li}.} \bibinfo{year}{2018}\natexlab{}.
\newblock \showarticletitle{Deep multi-view spatial-temporal network for taxi
  demand prediction}. In \bibinfo{booktitle}{\emph{Thirty-Second AAAI
  Conference on Artificial Intelligence}}.
\newblock


\bibitem[\protect\citeauthoryear{Yoo}{Yoo}{2019}]%
        {yoo2019short}
\bibfield{author}{\bibinfo{person}{Eun-hye Yoo}.}
  \bibinfo{year}{2019}\natexlab{}.
\newblock \showarticletitle{How short is long enough? Modeling temporal aspects
  of human mobility behavior using mobile phone data}.
\newblock \bibinfo{journal}{\emph{Annals of the American Association of
  Geographers}} \bibinfo{volume}{109}, \bibinfo{number}{5}
  (\bibinfo{year}{2019}), \bibinfo{pages}{1415--1432}.
\newblock


\bibitem[\protect\citeauthoryear{Yoo, Pu, Eum, and Jiang}{Yoo
  et~al\mbox{.}}{2021}]%
        {yoo2021impact}
\bibfield{author}{\bibinfo{person}{Eun-hye Yoo}, \bibinfo{person}{Qiang Pu},
  \bibinfo{person}{Youngseob Eum}, {and} \bibinfo{person}{Xiangyu Jiang}.}
  \bibinfo{year}{2021}\natexlab{}.
\newblock \showarticletitle{The Impact of Individual Mobility on Long-Term
  Exposure to Ambient PM2. 5: Assessing Effect Modification by Travel Patterns
  and Spatial Variability of PM2. 5}.
\newblock \bibinfo{journal}{\emph{International Journal of Environmental
  Research and Public Health}} \bibinfo{volume}{18}, \bibinfo{number}{4}
  (\bibinfo{year}{2021}), \bibinfo{pages}{2194}.
\newblock


\bibitem[\protect\citeauthoryear{Yoo, Roberts, Eum, and Shi}{Yoo
  et~al\mbox{.}}{2020}]%
        {yoo2020quality}
\bibfield{author}{\bibinfo{person}{Eun-Hye Yoo}, \bibinfo{person}{John~E
  Roberts}, \bibinfo{person}{Youngseob Eum}, {and} \bibinfo{person}{Youdi
  Shi}.} \bibinfo{year}{2020}\natexlab{}.
\newblock \showarticletitle{Quality of hybrid location data drawn from
  GPS-enabled mobile phones: Does it matter?}
\newblock \bibinfo{journal}{\emph{Transactions in GIS}} \bibinfo{volume}{24},
  \bibinfo{number}{2} (\bibinfo{year}{2020}), \bibinfo{pages}{462--482}.
\newblock


\bibitem[\protect\citeauthoryear{Yuan, Zhang, Zhang, Geng, Cong, and Han}{Yuan
  et~al\mbox{.}}{2017}]%
        {yuan2017pred}
\bibfield{author}{\bibinfo{person}{Quan Yuan}, \bibinfo{person}{Wei Zhang},
  \bibinfo{person}{Chao Zhang}, \bibinfo{person}{Xinhe Geng},
  \bibinfo{person}{Gao Cong}, {and} \bibinfo{person}{Jiawei Han}.}
  \bibinfo{year}{2017}\natexlab{}.
\newblock \showarticletitle{PRED: periodic region detection for mobility
  modeling of social media users}. In \bibinfo{booktitle}{\emph{Proceedings of
  the Tenth ACM International Conference on Web Search and Data Mining}}. ACM,
  \bibinfo{pages}{263--272}.
\newblock


\bibitem[\protect\citeauthoryear{Yuan, Zhou, and Yang}{Yuan
  et~al\mbox{.}}{2018}]%
        {yuan2018hetero}
\bibfield{author}{\bibinfo{person}{Zhuoning Yuan}, \bibinfo{person}{Xun Zhou},
  {and} \bibinfo{person}{Tianbao Yang}.} \bibinfo{year}{2018}\natexlab{}.
\newblock \showarticletitle{Hetero-convlstm: A deep learning approach to
  traffic accident prediction on heterogeneous spatio-temporal data}. In
  \bibinfo{booktitle}{\emph{Proceedings of the 24th ACM SIGKDD International
  Conference on Knowledge Discovery \& Data Mining}}. ACM,
  \bibinfo{pages}{984--992}.
\newblock


\bibitem[\protect\citeauthoryear{Zaidi, Chandola, and Yoo}{Zaidi
  et~al\mbox{.}}{2021}]%
        {zaidi2021dst}
\bibfield{author}{\bibinfo{person}{Syed Mohammed~Arshad Zaidi},
  \bibinfo{person}{Varun Chandola}, {and} \bibinfo{person}{Eun-Hye Yoo}.}
  \bibinfo{year}{2021}\natexlab{}.
\newblock \showarticletitle{DST-Predict: Predicting Individual Mobility
  Patterns From Mobile Phone GPS Data}.
\newblock \bibinfo{journal}{\emph{Ieee Access}}  \bibinfo{volume}{9}
  (\bibinfo{year}{2021}), \bibinfo{pages}{167592--167604}.
\newblock


\bibitem[\protect\citeauthoryear{Zhang, Zheng, and Qi}{Zhang
  et~al\mbox{.}}{2017}]%
        {zhang2017deep}
\bibfield{author}{\bibinfo{person}{Junbo Zhang}, \bibinfo{person}{Yu Zheng},
  {and} \bibinfo{person}{Dekang Qi}.} \bibinfo{year}{2017}\natexlab{}.
\newblock \showarticletitle{Deep spatio-temporal residual networks for citywide
  crowd flows prediction}. In \bibinfo{booktitle}{\emph{Thirty-First AAAI
  Conference on Artificial Intelligence}}.
\newblock


\bibitem[\protect\citeauthoryear{Zheng, Han, and Sun}{Zheng
  et~al\mbox{.}}{2018}]%
        {zheng2018survey}
\bibfield{author}{\bibinfo{person}{Xin Zheng}, \bibinfo{person}{Jialong Han},
  {and} \bibinfo{person}{Aixin Sun}.} \bibinfo{year}{2018}\natexlab{}.
\newblock \showarticletitle{A survey of location prediction on twitter}.
\newblock \bibinfo{journal}{\emph{IEEE Transactions on Knowledge and Data
  Engineering}} \bibinfo{volume}{30}, \bibinfo{number}{9}
  (\bibinfo{year}{2018}), \bibinfo{pages}{1652--1671}.
\newblock


\bibitem[\protect\citeauthoryear{Zheng, Xie, Ma, et~al\mbox{.}}{Zheng
  et~al\mbox{.}}{2010}]%
        {zheng2010geolife}
\bibfield{author}{\bibinfo{person}{Yu Zheng}, \bibinfo{person}{Xing Xie},
  \bibinfo{person}{Wei-Ying Ma}, {et~al\mbox{.}}}
  \bibinfo{year}{2010}\natexlab{}.
\newblock \showarticletitle{Geolife: A collaborative social networking service
  among user, location and trajectory.}
\newblock \bibinfo{journal}{\emph{IEEE Data Eng. Bull.}} \bibinfo{volume}{33},
  \bibinfo{number}{2} (\bibinfo{year}{2010}), \bibinfo{pages}{32--39}.
\newblock


\end{thebibliography}

% % %% If your work has an appendix, this is the place to put it.
\appendix

\section{Evaluation measure}
To evaluate the predictive power of the proposed model to correctly identify the visit locations for a given individual, we need evaluation metrics that can measure the following two aspects:
\begin{enumerate}
    \item {\bf Recall} - {\em What fraction of actual visits were correctly predicted by the model?}
    \item {\bf Precision} - {\em What fraction of the predicted visits corresponded to the actual visits made by the individual?}
\end{enumerate}
Mathematically, the two quantities can be calculated as follows. Consider a $(M \times N)$ test image matrix, ${\bf X}$, at a given spatial scale, and let $\widehat{\bf X}$ be the corresponding prediction matrix, obtained from the model. Note that we have dropped the time subscript, $t$, for clarity. The recall and precision are defined as:
\setcounter{equation}{0}
\begin{eqnarray}
\text{recall} & = & \frac{\sum_{i,j=1}^{M,N}\text{min}({\bf X}_{ij},\widehat{{\bf X}}_{ij})}{\sum_{i,j=1}^{M,N}{\bf X}_{ij}}\label{eqn:recall}\\
\text{precision} & = & \frac{\sum_{i,j=1}^{M,N}\text{min}({\bf X}_{ij},\widehat{{\bf X}}_{ij})}{\sum_{i,j=1}^{M,N}\widehat{\bf X}_{ij}}\label{eqn:precision}
\end{eqnarray}
Note that, for both recall and precision, the numerator is the same and counts the overlap between the true and predicted visit counts for each grid cell.\ In the paper study, we report the average recall and precision over all daily visit counts matrices in the test data set.

An issue with the recall and precision metrics, as defined in~\eqref{eqn:recall} and~\eqref{eqn:precision}, is that they are dependent on a spatial scale (i.e.\ resolution) at which the matrices are created.\ Clearly, the task of predicting visit counts at a coarser resolution is {\em easier} than predicting visit counts at a finer resolution, and the expected recall and precision values at a coarser resolution are higher than at finer resolution.\ Consequently, the results obtained at different scales are incomparable. This is a clear shortcoming in the present context, since we are interested in understanding the performance of the proposed model as a function of the spatial scale. To address this issue, we propose scale-invariant versions of the above defined recall and precision metrics.

We first calculate the recall and precision of a naive predictor, which simply distributes the total visits in ${\bf X}$ uniformly across all the grid cells. The output of the naive predictor, denoted as $\widetilde{\bf X}$, is calculated as:
\begin{equation}
    \widetilde{\bf X}_{ij} = \frac{\sum_{i,j=1}^{M,N}{\bf X}_{ij}}{M\times N}
\end{equation}
The base recall and precision for this naive predictor are defined as:
\begin{eqnarray}
\text{base\_recall} & = & \frac{\sum_{i,j=1}^{M,N}\text{min}({\bf X}_{ij},\widetilde{{\bf X}}_{ij})}{\sum_{i,j=1}^{M,N}{\bf X}_{ij}}\label{eqn:baserecall}\\
\text{base\_precision} & = & \frac{\sum_{i,j=1}^{M,N}\text{min}({\bf X}_{ij},\widetilde{{\bf X}}_{ij})}{\sum_{i,j=1}^{M,N}\widetilde{\bf X}_{ij}}\label{eqn:baseprecision}
\end{eqnarray}
One can verify that the values for the base\_recall and base\_precision metrics likely increase as the spatial scale becomes coarser, because the probability of placing a randomly assigned visit to a correct grid cell by the naive predictor is $\frac{1}{M\times N}$, which increases as the scale becomes coarser, i.e., $M$ and $N$ become smaller. We use the performance of the naive predictor to ``normalize'' the recall and precision of the proposed model as follows:
\begin{eqnarray}
\text{norm\_recall} & = & \text{recall} - \text{base\_recall}\label{eqn:normrecall}\\
\text{norm\_precision} & = & \text{precision} - \text{base\_precision}\label{eqn:normprecision}
\end{eqnarray}
Both the normalized recall and precision values are reported when comparing the performance of the proposed model across different scales.








\end{document}